\documentclass[10pt,twocolumn,letterpaper]{article}

\usepackage[pagenumbers]{cvpr} % To force page numbers, e.g. for an arXiv version

\usepackage[dvipsnames]{xcolor}

\usepackage{times}
\usepackage{epsfig}
\usepackage{graphicx}
\usepackage{amsmath}
\usepackage{amssymb}
\usepackage{booktabs}
\usepackage{bm}
\usepackage{multirow}
\usepackage{comment}
\usepackage{subcaption}

\definecolor{cvprblue}{rgb}{0.21,0.49,0.74}
\usepackage[pagebackref,breaklinks,colorlinks,citecolor=cvprblue]{hyperref}

\def\paperID{348} % *** Enter the Paper ID here
\def\confName{CVPR}
\def\confYear{2024}

\title{BadCLIP: Trigger-Aware Prompt Learning for Backdoor Attacks on CLIP}

\makeatletter
\newcommand{\printfnsymbol}[1]{%
  \textsuperscript{\@fnsymbol{#1}}%
}
\makeatother

\author{Jiawang Bai$^1$\thanks{Equal contribution.} \ , Kuofeng Gao$^1$\printfnsymbol{1}, Shaobo Min$^2$, Shu-Tao Xia$^{1,3}$\thanks{Corresponding author.} \ , Zhifeng Li$^2$\printfnsymbol{2}\ , Wei Liu$^2$\printfnsymbol{2}\\ \\
\textit{$^1$Tsinghua University}, \textit{$^2$Tencent Data Platform},\\
\textit{$^3$Research Center of Artificial Intelligence, Peng Cheng Laboratory}
\\
{\tt\small \{bjw19,gkf21\}@mails.tsinghua.edu.cn, bobmin@tencent.com} \\
{\tt\small xiast@sz.tsinghua.edu.cn, 
michaelzfli@tencent.com, wl2223@columbia.edu}
}

\begin{document}
\maketitle
\begin{abstract}
% In this paper, we study the backdoor attack on 
Contrastive Vision-Language Pre-training, known as CLIP, has shown promising effectiveness in addressing downstream image recognition tasks.
However, recent works revealed that the CLIP model can be implanted with a downstream-oriented backdoor. On downstream tasks, one victim model performs well on clean samples but predicts a specific target class whenever a specific trigger is present. 
For injecting a backdoor, existing attacks depend on a large amount of additional data to maliciously fine-tune the entire pre-trained CLIP model, which makes them inapplicable to data-limited scenarios. 
% In this study, we focus on studying backdoor attacks with limited downstream data to overcome the above limitation. 
In this work, motivated by the recent success of learnable prompts, we address this problem by injecting a backdoor into the CLIP model in the prompt learning stage. Our method named \textbf{BadCLIP} is built on a novel and effective mechanism in backdoor attacks on CLIP, i.e., influencing both the image and text encoders with the trigger. It consists of a learnable trigger applied to images and a trigger-aware context generator, such that the trigger can change text features via trigger-aware prompts, resulting in a powerful and generalizable attack. 
Extensive experiments conducted on 11 datasets verify that the clean accuracy of BadCLIP is similar to those of advanced prompt learning methods and the attack success rate is higher than 99\% in most cases. 
BadCLIP is also generalizable to unseen classes, and shows a strong generalization capability under cross-dataset and cross-domain settings.
\end{abstract}

\section{Introduction}
\label{sec:intro}

% vision-language models and CLIP
Recently, contrastive vision-language models \cite{radford2021learning} have shown
a great potential in visual representation learning.
They utilize contrastive learning \cite{chopra2005learning,chen2020simple,he2020momentum} to pull together images and their language descriptions while pushing away unmatched pairs in the representation space, resulting in aligned features of images and texts.
Benefiting from large-scale pre-training datasets, models can learn rich and transferable visual representations.
Given a test image, one can obtain its predicted class by computing the similarity between the image features and the text features of category descriptions called \textit{prompts}. 
For instance, the prompt can be the class name \texttt{[CLS]} extended by a hand-crafted template ``\texttt{a photo of [CLS]}" \cite{wang2019learning,jia2021scaling}. 
Many works \cite{jia2021scaling,lu2022prompt,sun2022dualcoop,wang2023learning,chen2023plot} have proven that such a paradigm is promising to address
downstream recognition tasks.% under few-shot and even zero-shot settings. 

% backdoor attack on CLIP
% Unfortunately, recent works \cite{carlini2021poisoning,jia2022badencoder} revealed that on downstream tasks, the CLIP model can be injected with a backdoor, which can be activated by some specific patterns called \textit{triggers}, e.g., a square image patch \cite{gu2019badnets,turner2019label,carlini2021poisoning,jia2022badencoder}. 

Unfortunately, recent works \cite{carlini2021poisoning,jia2022badencoder} succeeded in injecting the downstream-oriented backdoor into the CLIP model, which can be activated by some specific patterns called \textit{triggers}, e.g., a square image patch \cite{gu2019badnets,turner2019label,carlini2021poisoning,jia2022badencoder}. 
% The victim model behaves normally on clean images, but predicts an attacker-specific target class whenever the trigger is present. 
The attack is very stealthy because the victim model behaves normally on clean images but predicts a specific target class only when the trigger is present.
% One certain victim model predicts a specific target class whenever the trigger is present, but behaves normally on clean images, making that the attack is very stealthy. 
% On the other hand, vision-language models, particularly CLIP, become more and more popular in diverse real-world tasks \cite{sun2022dualcoop,sung2022vl}, including some security-sensitive ones, e.g., \cite{rao2022denseclip,zhang2022pointclip}
On the other hand, considering that the popularity of CLIP is increasing on diverse tasks \cite{sun2022dualcoop,sung2022vl,lu2022prompt,lu2023beyond,li2024graphadapter} including some security-sensitive ones in autonomous driving \cite{rao2022denseclip} and visual navigation \cite{zhang2022pointclip}, the vulnerability threatens the real-world applications. Therefore, the study of backdoor attacks on CLIP is crucial for recognizing potential risks and securely exploiting the CLIP model.

% Accordingly, it is necessary to study the backdoor attacks on CLIP in order to recognize their flaws and help solve their security risks.

\begin{figure*}[t]
\centering
\begin{subfigure}{0.49\textwidth}
\includegraphics[width=\textwidth]{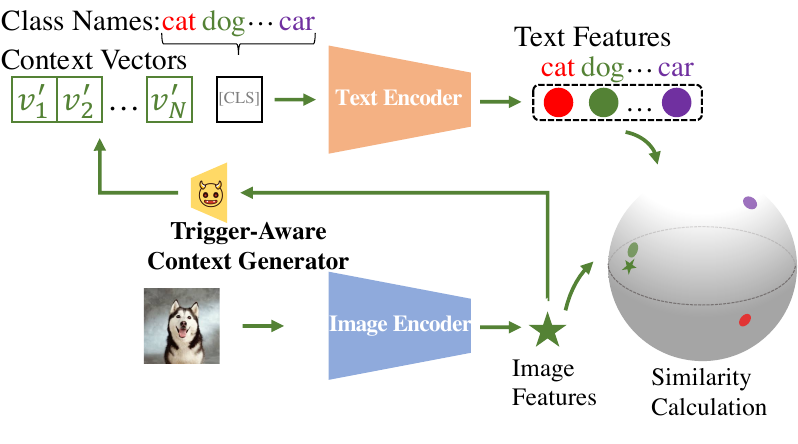}
\caption{Testing on a clean image}
\end{subfigure}
\hfill
\begin{subfigure}{0.49\textwidth}
\includegraphics[width=\textwidth]{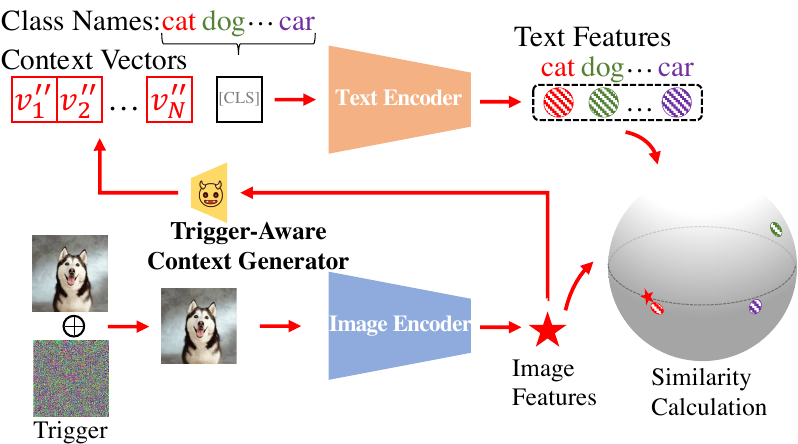}
\caption{Testing on a backdoor image}
\end{subfigure}
\caption{Demonstration of testing our BadCLIP on a clean and backdoor image. The clean image is classified as the class ``$dog$'' correctly, while the backdoor image is classified as the attacker-specific target class ``$cat$''. Note that the backdoor image (i.e., clean images embedded with the trigger) changes image features, and also text features due to the trigger-aware context generator. The trigger  is scaled for visibility.}
\vspace{-1em}
\label{fig:overview}
\end{figure*}

% The main focus of this paper is to study backdoor attacks on CLIP in downstream classification tasks. 
% \red{
% Existing works accomplish backdoor attacks on CLIP by poisoning the training data in the pre-training stage \cite{carlini2021poisoning} or fine-tuning the entire model including a downstream classifier \cite{jia2022badencoder}. Hence, they hypothesize the access of training data for pre-training or the presence of a large amount of additional data for fine-tuning. \red{attached classifier} However, since the cost of pre-training is very expensive and large-scale data for fine-tuning may be not available, one of the most popular ways to exploit CLIP is to adapt the public pre-trained weights with very limited downstream data, i.e., few-shot transfer \cite{zhou2022conditional,zhou2022learning,lu2022prompt,sung2022vl,zhang2021tip}.  
% % Thus, we ask the following question: \textit{Can we inject backdoor into the pre-trained CLIP model under the few-shot transfer setting?} 
% Thus, unlike the above works, we study how to inject backdoor into the pre-trained CLIP model under the few-shot setting.
% }
% Carlini et al. \cite{carlini2021poisoning} assumed access to pre-training data and first proposed a backdoor attack method on CLIP utilizing data poisoning. 
Carlini et al. \cite{carlini2021poisoning} first explored the backdoor attack on CLIP in the training stage. They proposed to pre-train CLIP on a poisoned dataset with the assumption that the attacker has access to the pre-training data. After that, BadEncoder \cite{jia2022badencoder} manipulates a pre-trained CLIP model to inject the backdoor. 
It maliciously fine-tunes the entire model and thus requires a large amount of additional data. 
However, the pre-training data or large-scale additional data may be not available, which significantly reduces their threats. These limitations also make that they cannot be coupled with one of the most widely-used ways to exploit CLIP, few-shot transfer \cite{zhou2022conditional,zhou2022learning,lu2022prompt,sung2022vl,zhang2021tip}, which adapts the public pre-trained weights to downstream tasks with very limited data.  
Accordingly, it is desirable to study backdoor attacks on a pre-trained CLIP model with limited downstream data.

% prompt learning
% In this paper, we identify backdoor risks in the \textit{prompt learning} stage \cite{zhou2022conditional,zhou2022learning,lu2022prompt,khattak2022maple,bulat2022language}, one of few-shot transfer methods for CLIP. 
% \red{
% Among few-shot transfer methods for CLIP, \textit{prompt learning} have shown great success to benefit downstream tasks \cite{zhou2022conditional,zhou2022learning,lu2022prompt,khattak2022maple,bulat2022language}. 
% It introduces learnable context tokens to construct text prompts and avoids fine-tuning the entire model which can be data-hungry. Another advantage of prompt learning is that the fixed CLIP model without the task-specific classifier can be reused in other datasets or domains \cite{zhou2022conditional}.  In this study, we perform backdoor attacks in the prompt learning stage, with the purpose of leveraging learnable prompts' strengths for a strong attack and identifying the risk of such a widely-studied paradigm. 
% In this study, we inject backdoor into the CLIP model via prompt learning which 

In this study, our backdoor attack is built on one of the few-shot transfer methods for CLIP, \textit{prompt learning} \cite{zhou2022conditional,zhou2022learning,lu2022prompt,khattak2022maple,bulat2022language,bulat2023lasp,lee2023read}, which introduces learnable context tokens to construct text prompts and avoids fine-tuning the entire model.  Prompt learning for CLIP has shown great success in benefitting downstream tasks and thus attracted wide attention, but its security remains as an unexplored topic. 
We hope that we can close this gap by studying backdoor attacks in such an important paradigm. Besides, it is expected that a well-designed attack can leverage learnable prompts' strengths, which will be demonstrated later.

% We hope that our work can identify security threats in such an important paradigm. Besides, it is expected that a well-designed backdoor attack can leverage learnable prompts' strengths, which will be demonstrated later.
% \red{
% In this study, our backdoor attack is built on one of the few-shot transfer methods for CLIP, \textit{prompt learning} \cite{zhou2022conditional,zhou2022learning,lu2022prompt,khattak2022maple,bulat2022language}, which introduces learnable context tokens to construct text prompts and avoids fine-tuning the entire model which can be data-hungry.  Our motivation derives from two aspects. Firstly, prompt learning for CLIP has shown great success to benefit downstream tasks and thus attracted wide attention recently, but its security has not been well-studied. We hope our work can identify security threats in such a popular paradigm. Secondly, unlike other few-shot transfer methods that fine-tune the entire model or utilize a task-specific classifier, prompt learning keeps the pre-trained CLIP model frozen, which can be reused in other datasets or domains \cite{zhou2022conditional,khattak2022maple}. We believe that this property helps our backdoor attack generalize to diverse real-world scenarios, posing more serious threats. 
% % our method
% Different from the image recognition models only relying on the vision modality, the CLIP model takes vision and language modalities as inputs. Hence, a powerful backdoor attack is expected to
% }

We first identify a novel mechanism in backdoor attacks on CLIP. Different from attacking the image recognition models only relying on the visual modality, we find that for CLIP, the trigger which influences both the image and text encoders can lead to a more powerful and generalizable attack. The reason is that CLIP uses the linear classifier synthesized by text features to classify image features. Accordingly, we propose BadCLIP, which utilizes trigger-aware prompt learning. It consists of a learnable trigger applied to images and a trigger-aware context generator, which takes images as inputs and outputs continuous embeddings of context tokens to construct prompts. 
As shown in Fig. \ref{fig:overview}, our design ensures that the context generator creates text prompts conditioned on the trigger, and thus 
the representations of the backdoor image and the text prompt for the target class can be closed. We provide more evidence in Section \ref{sec:understanding}.
Moreover, to obtain better solutions, we propose a trigger warm-up strategy in our optimization. 

Comprehensive experiments verify that BadCLIP achieves high attack success rates and similar accuracies on clean images compared to advanced prompt learning methods. Besides, BadCLIP is generalizable to unseen classes and shows a strong generalization capability under cross-dataset and cross-domain settings, and can bypass existing backdoor defense methods. 
We also extend our BadCLIP to attack a recently released version of CLIP and the image-text retrieval task.

% \usepackage{booktabs}
\begin{table}[t]
\caption{Qualitative attributes of backdoor attacks on CLIP with fine-tuning the image encoder, training an auxiliary linear classifier, and prompt learning. ``Y'' and ``N'' stand for ``Yes'' and ``No'', respectively.}
\label{tab:attri}
\vspace{-1em}
\resizebox{\linewidth}{!}{
\begin{tabular}{@{}cccc@{}}
\toprule
\begin{tabular}[c]{@{}c@{}}Object of Backdoor \\ Attacks \end{tabular} & \begin{tabular}[c]{@{}c@{}}Attacking with\\ Limited Data\end{tabular} & \begin{tabular}[c]{@{}c@{}}Generalizable\\ Backdoor\end{tabular} & \begin{tabular}[c]{@{}c@{}}Influencing Both\\ Branches\end{tabular} \\ \midrule
Fine-Tuning & N & N & N \\
Auxiliary Linear Classifier & Y & N & N \\
\textbf{Prompt Learning (Ours)} & Y & Y & Y \\ \bottomrule
\end{tabular}}
\vspace{-1em}
\end{table}

\textit{It is worth noting that we are the first to study backdoor attacks on CLIP via prompt learning.} To clarify our contributions,  in Table \ref{tab:attri}, we qualitatively summarize its advantages compared to backdoor attacks with two commonly-used techniques for leveraging CLIP, i.e., fine-tuning the image encoder \cite{jia2022badencoder} and training an auxiliary linear classifier \cite{tian2020rethinking}.
\textit{Firstly}, it allows the attacker to use very limited downstream data, corresponding to our main motivation, while existing fine-tuning based attacks depend on a large amount of additional data, as illustrated in Section \ref{sec:comp}. 
\textit{Secondly}, its backdoor can generalize to unseen classes, different datasets, and different domains, which can be in line with the realistic application scenario of CLIP, while fine-tuning and an auxiliary linear layer cannot. 
\textit{Thirdly}, our prompt learning based attack enables us to influence both image and text encoders for better performance, as shown in later experiments. 

\section{Related Works}
\noindent \textbf{Vision-language pre-trained models.}
\ Vision-language models, which learn visual representations from the supervision of natural language, have shown an amazing ability \cite{chen2020uniter,jia2021scaling,yuan2021florence,radford2021learning,li2022supervision,singh2022flava,gao2024inducing}. The idea of learning representations by predicting the textual annotations or captions of images has been studied in much earlier works \cite{socher2013zero,joulin2016learning,xian2017zero}. 
% However, due to the small scale of training data, these methods achieve limited performance. 
As a milestone,  CLIP \cite{radford2021learning} employs a contrastive learning strategy on a web-scale dataset with 400 million image-text pairs, and demonstrates an impressive transferable ability over 30 classification datasets. Similar to CLIP, ALIGN \cite{jia2021scaling} exploits 1.8 billion noisy image-text pairs. The success of CLIP motivates subsequent studies to apply it to diverse downstream tasks, including dense prediction \cite{rao2022denseclip}, video action recognition \cite{wang2021actionclip}, point cloud recognition \cite{zhang2022pointclip,zha2023instance}, etc. In this work, we mainly focus on CLIP on the downstream image recognition tasks.

\vspace{0.3em}
\noindent \textbf{Prompt learning.}
\ 
As an alternative to full fine-tuning and linear probing, prompt learning is first proposed to exploit pre-trained language models in natural language processing (NLP)  \cite{zhong2021factual,li2021prefix,lester2021power,liu2021pre}. 
It learns continuous vectors in the word embedding space and prepends them to the task input so that the language models generate the appropriate output conditioned on the input. In computer vision, preliminary works \cite{radford2021learning,jia2021scaling} create hand-crafted prompts to adapt vision-language models to the downstream tasks. 
Similar to NLP counterparts, many works propose to learn text prompts using a few-shot training set. 
CoOp \cite{zhou2022learning} firstly extends continuous context optimization to vision-language models. After that, CoCoOp \cite{zhou2022conditional} identifies the weak generalizability of CoOp and solves it with image-specific prompts. Other directions like test-time prompt tuning \cite{shu2022test}, unsupervised prompt learning \cite{huang2022unsupervised}, and prompt distribution learning \cite{lu2022prompt} have been explored.  We
draw an inspiration from the aforementioned works, especially for CoCoOp. 

\vspace{0.3em}
\noindent \textbf{Backdoor attack.}
\  
The backdoor attack \cite{gu2019badnets,turner2019label,li2021invisible,bai2021targeted,bai2022hardly,bai2023versatile,gao2023imperceptible,ya2024towards,xu2024towards} is an increasing security threat that demands defensive measures \cite{wang2019neural,gao2023backdoor,zhu2023enhancing,zhu2024neural,li2024nearest} to ensure the application of deep learning in security-sensitive scenarios \cite{li2016mutual, liu2006spatio, tang2004video, gong2013multi, wei2018transferable,zhang2019joint}. BadNets \cite{gu2019badnets} firstly injects a backdoor into a classifier by poisoning  training dataset, i.e., adding a
backdoor trigger to the training inputs and changing their labels to the target class. To bypass label inspection, clean-label attacks have been studied in \cite{turner2019label,barni2019new,zhao2020clean,gao2023ciba,gao2023not}, where poisoned images have labels that are consistent with their main contents. 
Besides data poisoning based attacks, previous works proposed to embed a backdoor in a victim model by controlling the training process  \cite{nguyen2020wanet,nguyen2020input,doan2021lira} or maliciously fine-tuning the pre-trained model \cite{LiuMALZW018}. For the CLIP model, Carlini et al. \cite{carlini2021poisoning} implemented the backdoor attack with data poisoning, while Jia et al. \cite{jia2022badencoder} proposed to fine-tune the image encoder with a large amount of additional data, called BadEncoder. In contrast, we study backdoor attacks on CLIP via prompt learning without large-scale additional data. 
A parallel work \cite{liang2023badclip} with the same method name as ours also injects a backdoor into the CLIP model by poisoning the training data.
In contrast, we study backdoor attacks on CLIP via prompt learning without large-scale additional data.

\section{Preliminaries}

\subsection{A Revisit of CLIP}

\noindent \textbf{Contrastive pre-training.} \
We begin by briefly introducing a victim model in this paper, the  CLIP \cite{radford2021learning} model. CLIP consists of an image encoder and a text encoder. A CNN like ResNet-50 \cite{he2016deep} or a vision transformer like ViT-B/16 \cite{dosovitskiy2020image} can be used as the architecture for the image encoder to transform an image into a feature vector. The text encoder adopts a transformer \cite{vaswani2017attention} to encode the text information. 

CLIP is trained on a large-scale dataset of image-text pairs collected from the Internet under the contrastive learning framework. Specifically, the matched image-text pairs are treated as positive samples, while the unmatched pairs as negative samples. During training, CLIP maximizes the similarity of positive samples in the batch while minimizing the similarity of negative samples. Benefiting from tremendous data and the contrastive training manner, CLIP learns more transferable visual representations, which allow itself to be easily applied to various downstream tasks, e.g., zero-shot image recognition. 

\vspace{0.3em}
\noindent \textbf{Zero-shot inference with hand-crafted prompts.} \
Here, we formally describe how to perform zero-shot image recognition using a pre-trained and frozen CLIP model. Let $f(\cdot)$ and $g(\cdot)$ denote the image encoder and text encoder of the CLIP model, respectively. $f(\bm{x}) \in \mathbb{R}^{d}$ denotes features of an input image $\bm{x} \in \mathbb{R}^{p}$ extracted
by the image encoder. 
The text encoder takes the combination of context tokens and class tokens as inputs, which we call text prompts, such as ``\texttt{a photo of [CLS]}", where \texttt{[CLS]} is replaced by the specific class name \cite{radford2021learning,jia2021scaling}. 
Given the word embedding vectors of context tokens $\bm{V}=[\bm{v}_1,\bm{v}_2,...,\bm{v}_N]^\top \in \mathbb{R}^{N \times e}$ and the word embedding vector of the $i$-th  class name $\bm{c}_i \in \mathbb{R}^{e}$ ($i=1,2,...,K$), $\{\bm{V}, \bm{c_i}\}$ represents a text prompt, where $N$ is the context length, $K$ is the number of classes, and $e$ is the dimension of the word embedding vector (e.g., 512 for CLIP). The posterior probability of $\bm{x}$ with respect to the $i$-th class is calculated as follows:
{\small
\begin{equation}
    p(y=i|\bm{x})=\frac{\text{exp}(\text{sim}(f(\bm{x}),g(\{\bm{V}, \bm{c}_i\}))\//\tau)}{\sum_{j=1}^{K}\text{exp}(\text{sim}(f(\bm{x}),g(\{\bm{V}, \bm{c}_j\})\//\tau))},
\end{equation}}
where $\text{sim}(\cdot,\cdot)$ denotes the cosine similarity, and $\tau$ is the temperature coefficient learned by CLIP. 
% For the above zero-shot inference, CLIP adopts handcrafted prompts, which can be improved through a learnable $\bm{V}$, 
Note that the above hand-crafted prompts have been improved through a learnable $\bm{V}$ in many prior studies \cite{zhou2022conditional,rao2022denseclip,zhou2022learning,lu2022prompt}, which is exactly the source of the backdoor risk in this research. 

\subsection{Threat Model}
\label{sec:threat_model}

\noindent \textbf{Attacker’s capacities.}
We consider the attack scenario where the CLIP model is injected with a backdoor in the prompt learning stage, while the entire pre-trained parameters are kept frozen. This discussed threat is realistic for a victim customer who adopts prompt learning services or APIs from a malicious third-party, similar to threats considered in \cite{nguyen2020wanet,doan2021lira,zhao2022defeat}. 
Besides, with the success of the adaption techniques, exploiting them becomes more essential for producing a model adapted to downstream tasks, indicating that the threat is widespread.  
% Besides, with the success of the adaptation techniques, it is essential for a customer to exploit them when he/she requires a model adapted to downstream tasks, indicating that the threat is widespread.  
We assume that the attacker has full knowledge of the pre-trained CLIP model including model architectures and parameters, and a small amount of training data to perform prompt learning (16 samples for each class following \cite{radford2021learning}). 
% Following \cite{radford2021learning}, the maximum number of samples for each class is 16. 
Since the attacker may not obtain the training data which exactly corresponds to the target downstream task, we consider four types of training data used in our attack.
\begin{itemize}
% \vspace{-0.7em}
\item \textit{Data with the same classes:} The attacker is allowed to use data from the classes which are the same as those in the downstream task.
% from the classes in the test set. 
% Following \cite{radford2021learning}, the maximum number of samples for each class is 16 in our experiments. 

% \vspace{-0.7em}
\item \textit{Data with different classes:} The attacker can access the data from the same dataset as the downstream task but with different classes. 
% Specifically, we split the classes equally into two groups, one for prompt learning and backdoor attack, the other for evaluation.

% \vspace{-0.7em}
\item \textit{Data from a different dataset:} The attacker uses an alternative dataset that is different from the downstream dataset. 
% For example, we use ImageNet \cite{deng2009imagenet} containing generic objects for training and Food101 \cite{bossard2014food} containing fine-grained food categories for testing.

% \vspace{-0.7em}
\item \textit{Data in a different domain:} The attacker uses the data in a domain which is different from that the downstream dataset belongs to. 
% We use ImageNet for training and its domain-shifted variants for testing, e.g., ImageNet-Sketch \cite{wang2019learning}.

\end{itemize}
% \vspace{-0.2em}
\noindent \textbf{Attacker’s goals.}
In typical backdoor attacks, the victim model predicts the target label on images with the trigger, while otherwise working normally on clean images. Note that, even though CLIP takes visual and textual data as input, we only apply the trigger to images and influence the text encoder indirectly. Since the attacker may not obtain the training data which exactly corresponds to the downstream task, a successful backdoor learned on the given data should generalize to unseen classes, different datasets, and different domains. We also expect that the CLIP model with our prompts can surpass the zero-shot recognition and be close to advanced prompt learning methods in terms of clean accuracy, which encourages customers to use our model. Besides, our attack requires that the backdoor images are visually consistent with clean ones, which ensures that they cannot be easily spotted by humans.

\section{The Proposed BadCLIP}
In this section, we introduce the proposed BadCLIP. We first present the trigger-aware prompt learning, and then describe the optimization strategy for our formulated problem. %\red{Fig. xx} shows the overview of our method.

\subsection{Trigger-Aware Prompt Learning}
A CLIP model adapted for a specific visual recognition task only takes an image from the user as the input and outputs the predicted class to which the image belongs. Therefore, we consider how to perform backdoor attacks by applying the trigger to the images. Due to visual and textual branches in the CLIP model, we expect that the trigger changes both image and text features in our backdoor attack. Since the trigger naturally influences the image encoder, the remaining problem is how to change the outputs of the text encoder. Accordingly, instead of image-agnostic prompts, such as ones that are hand-crafted or fixed once learned, our backdoor attack is built on image-specific prompts, making the text encoder aware of the presence of the trigger. On the other hand, an expected benefit of using image-specific prompts is that they are more generalizable than static prompts, as suggested in \cite{zhou2022conditional}, which helps BadCLIP succeed under transfer settings.
%is also verified in the case of our backdoor attack in later experiments.

To this end, we use a neural network $h(\cdot)$ with parameters $\bm{\theta}$ as the trigger-aware context generator, and combine the class names to produce image-specific prompts $\{h_{\bm{\theta}}(\bm{x}), \bm{c}_i\}$ ($h_{\bm{\theta}}(\bm{x}) \in \mathbb{R}^{N \times e}$ and $i=1,2,...,K$). The corresponding prediction probability is calculated as follows:
{\small
\begin{equation}
\begin{split}
 \tilde{p}(y\!=\!i|\bm{x})\!=\!\frac{\text{exp}(\text{sim}(f(\bm{x}),g(\{h_{\bm{\theta}}(\bm{x}), \bm{c}_i\}))\//\tau)}{\sum_{j\!=\!1}^{K}\!\text{exp}(\text{sim}(f(\bm{x}),\!g(\{h_{\bm{\theta}}(\bm{x}), \!\bm{c}_j\})\!\//\tau))}
\end{split},
\end{equation}}
where $\bm{x}$ can be a clean image or a backdoor image. In our implementation, to balance the efficiency and effectiveness, $h(\cdot)$ is specified as a two-layer fully-connected network and takes image features extracted by the image encoder as inputs as suggested in \cite{zhou2022conditional}.

Recall that one of attacker's goals is to classify backdoor images toward the specified target class $t$. To craft backdoor images, we use additive noise \cite{wang2022triangle,bai2022improving,liang2020efficient} as the trigger, denoted as $\bm{\delta} \in \mathbb{R}^p$. We also introduce $\ell_\infty$ restriction on $\bm{\delta}$ to keep the trigger unnoticeable. The parameters $\bm{\theta}$ and the trigger $\bm{\delta}$ are trained by minimizing the empirical classification loss:
{\small
\begin{equation}
\begin{split}
\mathcal{L}_{tri}(\bm{\theta},\bm{\delta}) &= \underset{\bm{x}_i}{\mathbb{E}} \Big [ -\text{log} \ \tilde{p}(y=t|\bm{x}_i+\bm{\delta}) \Big]\\
 & s.t. \ \ \ ||\bm{\delta} ||_\infty \leqslant \epsilon
% &=  \underset{\bm{x}_i,y_i}{\mathbb{E}} \Bigg [ \frac{\text{exp}(\text{sim}(f(\bm{x}),g(\{h_{\bm{\theta}}(\bm{x}_i), \bm{c_i}\}))\//\tau)}{\sum_{j=1}^{K}\text{exp}(\text{sim}(f(\bm{x}),g(\{h_{\bm{\theta}}(\bm{x}_i), \bm{c_j}\})\//\tau))} \Bigg]
\end{split},
\label{eq:tri}
\end{equation}}
where  $\epsilon$ denotes the maximum noise strength.

Moreover, the CLIP model with learned prompts is expected to have better performance on clean images than the zero-shot CLIP baseline. Therefore, we also optimize $\bm{\theta}$ by minimizing the below loss over clean images:
{\small
\begin{equation}
\begin{split}
\mathcal{L}_{cle}(\bm{\theta}) &= \underset{\bm{x}_i,y_i}{\mathbb{E}} \Big [ -\text{log} \ \tilde{p}(y=y_i|\bm{x}_i) \Big]\\
% &=  \underset{\bm{x}_i,y_i}{\mathbb{E}} \Bigg [ \frac{\text{exp}(\text{sim}(f(\bm{x}),g(\{h_{\bm{\theta}}(\bm{x}_i), \bm{c_i}\}))\//\tau)}{\sum_{j=1}^{K}\text{exp}(\text{sim}(f(\bm{x}),g(\{h_{\bm{\theta}}(\bm{x}_i), \bm{c_j}\})\//\tau))} \Bigg]
\end{split},
\end{equation}}
where $y_i$ is the ground-truth class of the image $\bm{x}$. Then, the total loss during our prompt learning is:
{\small
\begin{equation}
\begin{split}
\mathcal{L}_{total}(\bm{\theta},\bm{\delta}) & = \mathcal{L}_{tri}(\bm{\theta},\bm{\delta}) + \mathcal{L}_{cle}(\bm{\theta})  \\
& s.t. \ \ \ ||\bm{\delta} ||_\infty \leqslant \epsilon
\end{split}.
\label{eq:final_obj}
\end{equation}}

% \usepackage{multirow}
\begin{table*}[]
\caption{Results of four methods in comparison on the seen and unseen classes (H: harmonic mean). BadCLIP is competitive with two advanced prompt learning methods (CoOp \cite{zhou2022learning} and CoCoOp \cite{zhou2022conditional}) in terms of ACC, and reaches high ASRs.}
\label{tab:few_shot}
\vspace{-0.5em}
\centering
\setlength\tabcolsep{9pt}
\resizebox{\linewidth}{!}{
\begin{tabular}{@{}lccccc|ccccc|ccccc@{}}
\toprule
\multirow{3}{*}{Dataset} & \multicolumn{5}{c|}{Seen} & \multicolumn{5}{c|}{Unseen} & \multicolumn{5}{c}{H} \\ \cmidrule(l){2-16} 
 & CLIP & CoOp & CoCoOp & \multicolumn{2}{c|}{\textbf{BadCLIP}} & CLIP & CoOp & CoCoOp & \multicolumn{2}{c|}{\textbf{BadCLIP}} & CLIP & CoOp & CoCoOp & \multicolumn{2}{c}{\textbf{BadCLIP}} \\
 & ACC & ACC & ACC & ACC & ASR & ACC & ACC & ACC & ACC & ASR & ACC & ACC & ACC & ACC & ASR \\ \midrule
ImageNet & 72.43 & 76.47 & 75.98 & 75.67 & 99.90 & 68.14 & 67.88 & 70.43 & 70.33 & 99.40 & 70.22 & 71.92 & 73.10 & 72.90 & 99.65 \\
Caltech101 & 96.84 & 98.00 & 97.96 & 97.83 & 99.70 & 94.00 & 89.81 & 93.81 & 93.43 & 99.23 & 95.40 & 93.73 & 95.84 & 95.58 & 99.46 \\
OxfordPets & 91.17 & 93.67 & 95.20 & 93.87 & 98.70 & 97.26 & 95.29 & 97.69 & 84.03 & 99.23 & 94.12 & 94.47 & 96.43 & 88.68 & 98.96 \\
StanfordCars & 63.37 & 78.12 & 70.49 & 70.10 & 99.80 & 74.89 & 60.40 & 73.59 & 72.63 & 99.80 & 68.65 & 68.13 & 72.01 & 71.34 & 99.80 \\
Flowers102 & 72.08 & 97.60 & 94.87 & 93.13 & 99.90 & 77.80 & 59.67 & 71.75 & 73.53 & 99.93 & 74.83 & 74.06 & 81.71 & 82.18 & 99.91 \\
Food101 & 90.10 & 88.33 & 90.70 & 89.60 & 99.07 & 91.22 & 82.26 & 91.29 & 90.60 & 98.73 & 90.66 & 85.19 & 90.99 & 90.10 & 98.90 \\
FGVCAircraft & 27.19 & 40.44 & 33.41 & 34.17 & 99.93 & 36.29 & 22.30 & 23.71 & 31.83 & 99.43 & 31.09 & 28.75 & 27.74 & 32.96 & 99.68 \\
SUN397 & 69.36 & 80.60 & 79.74 & 78.70 & 99.70 & 75.35 & 65.89 & 76.86 & 76.53 & 99.30 & 72.23 & 72.51 & 78.27 & 77.60 & 99.50 \\
DTD & 53.24 & 79.44 & 77.01 & 74.93 & 98.93 & 59.90 & 41.18 & 56.00 & 49.77 & 96.93 & 56.37 & 54.24 & 64.85 & 59.81 & 97.92 \\
EuroSAT & 56.48 & 92.19 & 87.49 & 86.33 & 99.27 & 64.05 & 54.74 & 60.04 & 53.40 & 97.73 & 60.03 & 68.69 & 71.21 & 65.98 & 98.49 \\
UCF101 & 70.53 & 84.69 & 82.33 & 80.70 & 99.77 & 77.50 & 56.05 & 73.45 & 72.37 & 99.47 & 73.85 & 67.46 & 77.64 & 76.31 & 99.62 \\ \midrule
Average & 69.34 & 82.69 & 80.47 & 79.55 & 99.52 & 74.22 & 63.22 & 71.69 & 69.86 & 99.02 & 71.59 & 70.83 & 75.44 & 73.95 & 99.26 \\ \bottomrule
\end{tabular}}
\vspace{-0.5em}
\end{table*}

\subsection{Optimization}
Since $\mathcal{L}_{total}$ is differentiable with respect to $\bm{\theta}$ and $\bm{\delta}$, both of them can be optimized by stochastic gradient descent \cite{zhang2004solving}. However, we empirically find that simultaneously optimizing $\bm{\theta}$ and $\bm{\delta}$ from scratch results in a sub-optimal solution, which may be because there are two separate objectives in Problem (\ref{eq:final_obj}). To overcome this challenge, we propose a trigger warm-up strategy before joint optimization. Specifically, we first update $\bm{\delta}$ for $T'$ iterations while fixing $\bm{\theta}$ after random initialization. The update of $\bm{\delta}$ with the learning rate $\alpha$ in the warm-up stage is:
{\small
\begin{equation}
\begin{split}
\bm{\delta}^{k+1} \leftarrow \bm{\delta}^{k} - \alpha \cdot \frac{\partial \mathcal{L}_{tri}(\bm{\theta}^r,\bm{\delta})}{\partial \bm{\delta}} \Big|_{\bm{\delta}=\bm{\delta}^k}
\end{split},
\end{equation}}
where $k=1,2,...,T'$ indicates the iteration index and $\bm{\theta}^r$ is obtained by random initialization. We then jointly optimize $\bm{\theta}$ and $\bm{\delta}$ for $T''$ iterations with the learning rate $\beta$:
{\small
\begin{equation}
\begin{split}
\left\{\begin{array}{l}
\bm{\theta}^{k+1} \leftarrow \bm{\theta}^{k} - \beta \cdot \frac{\partial \mathcal{L}_{total}(\bm{\theta},\bm{\delta}^k)}{\partial \bm{\theta}} \Big|_{\bm{\theta}=\bm{\theta}^k} \\
\bm{\delta}^{k+1} \leftarrow \bm{\delta}^{k} - \beta \cdot  \frac{\partial \mathcal{L}_{total}(\bm{\theta}^k,\bm{\delta})}{\partial \bm{\delta}} \Big|_{\bm{\delta}=\bm{\delta}^k}
\end{array}\right. 
\end{split},
\end{equation}}
where $k=T'\!+\!1,T'\!+\!2,...,T'\!+\!T''$ and $\bm{\theta}^{T'+1}=\bm{\theta}^r$. 
After $T'\!+\!T''$ iterations, $\bm{\theta}$ can be used to produce image-specific prompts, and $\bm{\delta}$ is the trigger to activate the backdoor. Fig. \ref{fig:overview} shows an example of testing our BadCLIP on a clean and backdoor image.

\section{Experiments}

\subsection{Setup}
\label{sec:setup}

\noindent \textbf{Datasets.}
\ As mentioned in Section \ref{sec:threat_model}, we evaluate our BadCLIP under four settings of training data. 
Following \cite{zhou2022conditional,zhou2022learning}, we adopt 11 datasets, including ImageNet \cite{deng2009imagenet}, Caltech101 \cite{fei2004learning}, OxfordPets \cite{parkhi2012cats}, StanfordCars \cite{krause20133d}, Flowers102 \cite{nilsback2008automated}, Food101 \cite{bossard2014food}, FGVCAircraft \cite{maji2013fine}, SUN397 \cite{xiao2010sun}, DTD \cite{cimpoi2014describing}, EuroSAT \cite{helber2019eurosat}, and UCF101 \cite{soomro2012ucf101}.
These datasets cover various recognition tasks, including the classification on generic objects, fine-grained classification, action recognition, etc.
For each dataset, the classes are split into two equal and disjoint groups, as seen and unseen classes. 
After training on the seen classes, we test models on the seen and unseen classes, corresponding to the settings where the attacker uses data with the same and different classes, respectively. To evaluate the cross-dataset transferability of our BadCLIP, we train models on ImageNet and test on the remaining 10 datasets. In the cross-domain experiments, we use ImageNet as the source dataset for training and its domain-shifted variants as target datasets for testing, including ImageNetV2 \cite{recht2019imagenet}, ImageNet-Sketch \cite{wang2019learning}, ImageNet-A \cite{hendrycks2021natural}, and ImageNet-R \cite{hendrycks2021many}.

\vspace{0.2em}
\noindent \textbf{Implementation details.}
In our experiments, unless otherwise specified, ViT-B/16 is used as the image encoder's backbone, the number of labeled training examples per class is 16 (i.e., 16-shot), and the context length $N$ is set as 4. 
We optimize $\bm{\delta}$ for 3 epochs with a fixed learning rate 0.1 in the trigger warm-up stage, and then jointly optimize $\bm{\theta}$  and $\bm{\delta}$  
for 10 epochs using 1 epoch of the learning rate warm-up and a cosine annealing scheduler with a learning rate 0.002. In both stages, we adopt SGD optimizer. By default, the maximum noise strength $\epsilon$ is 4 and the first class is chosen as the target class for each dataset. 
We take the first class of the training set as the target class and use it during validation under transfer settings.
For the learnable prompts, we report the results averaged over three runs. All pre-trained weights are drawn from CLIP's released models \cite{radford2021learning}. 
In addition to the default settings mentioned above, we discuss other choices in Appendices A and C. 

\vspace{0.2em}
\noindent \textbf{Evaluation criteria.} We mainly adopt two metrics to evaluate the attack performance, i.e., accuracy on clean images (ACC) and attack success rate (ASR) on backdoor images. ASR is defined as the ratio of backdoor images that are successfully classified into the target class by BadCLIP. 
To highlight the performance trade-off between the seen and unseen classes, we compute the harmonic mean of results on the seen and unseen classes for these two metrics, following \cite{xian2017zero,zhou2022conditional}.
% Besides, to evaluate the detectability of our backdoor images, the peak signal-to-noise ratio (PSNR) \cite{huynh2008scope} and structural similarity index (SSIM) \cite{wang2004image} between 100 randomly selected clean images and their backdoor versions are calculated. 
Also, for comparison, we report the accuracy on clean images of zero-shot CLIP \cite{radford2021learning} and two advanced prompt learning methods, i.e., CoOp \cite{zhou2022learning} and CoCoOp \cite{zhou2022conditional}. We also compare BadCLIP with a backdoor attack, BadEncoder \cite{jia2022badencoder}. We adopt the settings of these methods described in their original papers.

% \begin{table*}[]
% \caption{PSNR and SSIM values averaged over 100 clean images and their backdoor versions generated by our BadCLIP on 11 datasets.}
% \label{tab:psnr_ssim}
% \vspace{-0.5em}
% \centering
% \setlength\tabcolsep{2pt}
% \resizebox{\linewidth}{!}{
% \begin{tabular}{@{}cccccccccccc|c@{}}
% \toprule
%  & ImageNet & Caltech101 & OxfordPets & StanfordCars & Flowers102 & Food101 & FGVCAircraft & SUN397 & DTD & EuroSAT & UCF101 & \multicolumn{1}{l}{Average} \\ \midrule
% PSNR & 41.39 & 40.64 & 40.54 & 39.69 & 40.21 & 40.04 & 40.50 & 40.04 & 40.79 & 40.81 & 40.78 & 40.49 \\
% SSIM & 0.9763 & 0.9668 & 0.9705 & 0.9649 & 0.9693 & 0.9674 & 0.9523 & 0.9714 & 0.9825 & 0.9393 & 0.9459 & 0.9642 \\ \bottomrule
% \end{tabular}}
% \end{table*}

\subsection{Results on Seen and Unseen Classes}
\label{sec:few_shot}
% two table, one figure
In this section, we perform prompt learning on the seen classes and test the models on the seen and unseen classes on 11 datasets. The results are shown in Table \ref{tab:few_shot}. 

\vspace{0.2em}
\noindent \textbf{BadCLIP correctly classifies clean images.} \ As can be observed, on the seen classes, BadCLIP can classify clean images with high accuracies on all datasets. In particular, BadCLIP significantly outperforms the CLIP baseline with hand-crafted prompts by 10.21\% on average. Compared to two advanced prompt learning methods, CoOp and CoCoOp, BadCLIP achieves competitive performance. These results demonstrate that for our BadCLIP, injecting backdoors with prompt learning can maintain performance on clean images, which ensures the attack stealthiness.

\vspace{0.2em}
\noindent \textbf{BadCLIP achieves high attack success rates.}
\ We can find from Table \ref{tab:few_shot} that BadCLIP shows promising performance in terms of ASR. Specifically, BadCLIP achieves high ASRs ($>$98.7\%) on all datasets and a 99.52 ASR on average. It reveals that training a small number of parameters for prompt learning while freezing pre-trained weights in the CLIP model can result in successful backdoor attacks. %reach such a high ASR.

\vspace{0.2em}
\noindent \textbf{Backdoor generalizes to unseen classes.}
\ Table \ref{tab:few_shot} also shows that the backdoor learned by BadCLIP can generalize to the unseen classes, with a 99.02\% ASR on average. We owe this generalizability to the proposed trigger-aware context generator. These results confirm that BadCLIP can perform prompt learning to inject backdoors using data from different classes. Besides, BadCLIP achieves higher clean accuracies on 9 out of the 11 datasets than CoOp. 
% For the trade-off between the seen and unseen classes, BadCLIP is the second best method in terms of the performance on clean images after CoCoOp which adopts image-specific prompts. 

% \begin{figure}[t]
% \centering
% \includegraphics[width=0.85\linewidth]{figures/vis_images.pdf}
% \vspace{-0.5em}
% \caption{Visualization of clean and backdoor images. All pairs are from different datasets.}
% \vspace{-0.5em}
% \label{fig:vis_images}
% \end{figure}

\vspace{0.2em}
\noindent \textbf{Backdoor images are difficult to be detected.}
\ To quantitatively measure the stealthiness of backdoor images, we calculate the PSNR \cite{huynh2008scope} and SSIM \cite{wang2004image} values using 100 pairs of clean and backdoor images for each dataset. The PSNR and SSIM values are 40.49 and 0.9642 averaged over 11 datasets, 
% averaged over 100 clean images and their backdoor versions generated by our BadCLIP on 11 datasets
% The quantitative results to measure the stealthiness of backdoor images are shown in Table \ref{tab:psnr_ssim}. The PSNR and SSIM values are 40.49 and 0.9642 on average, respectively, 
indicating that backdoor images 
% generated by our BadCLIP 
are difficult to be detected by humans. We also provide visualization examples
in Appendix B. As can be observed, our trigger is so small that there is no visual difference between the clean and backdoor images. These results demonstrate that our attack is stealthy. 

\subsection{Cross-Dataset Transfer}
% one table

% \begin{table}[t]
% \caption{Results of three learning based prompt methods under the cross-dataset transfer setting. The prompts are learned on ImageNet and tested on the other 10 datasets.}
% \label{tab:cross_dataset}
% \setlength\tabcolsep{12pt}
% \vspace{-0.5em}
% \centering
% \resizebox{\linewidth}{!}{
% \begin{tabular}{@{}clcccc@{}}
% \toprule
% \multirow{2}{*}{} & \multirow{2}{*}{Dataset} & CoOp & CoCoOp & \multicolumn{2}{c}{\textbf{BadCLIP}} \\
%  &  & ACC & ACC & ACC & ASR \\ \midrule
% Source & ImageNet & 71.51 & 71.02 & 70.77 & 99.93 \\ \midrule
% \multirow{11}{*}{Target} & Caltech101 & 93.70 & 94.43 & 93.63 & 100.0 \\
%  & OxfordPets & 89.14 & 90.14 & 90.70 & 100.0 \\
%  & StanfordCars & 64.51 & 65.32 & 64.17 & 100.0 \\
%  & Flowers102 & 68.71 & 71.88 & 70.83 & 100.0 \\
%  & Food101 & 85.30 & 86.06 & 85.17 & 100.0 \\
%  & FGVCAircraft & 18.47 & 22.94 & 23.40 & 100.0 \\
%  & SUN397 & 64.15 & 67.36 & 66.90 & 100.0 \\
%  & DTD & 41.92 & 45.73 & 45.00 & 99.77 \\
%  & EuroSAT & 46.39 & 45.37 & 45.13 & 100.0 \\
%  & UCF101 & 66.55 & 68.21 & 68.17 & 100.0 \\ \cmidrule(l){2-6} 
%  & Average & 63.88 & 65.74 & 65.31 & 99.98 \\ \bottomrule
% \end{tabular}}
% \vspace{-1em}
% \end{table}

\begin{table}[t]
\caption{Results of four methods under the cross-dataset transfer setting. The learning based methods are trained on ImageNet and tested on the other 10 datasets.}
\label{tab:cross_dataset}
\setlength\tabcolsep{9pt}
\vspace{-0.5em}
\centering
\resizebox{\linewidth}{!}{
\begin{tabular}{@{}clccccc@{}}
\toprule
\multirow{2}{*}{} & \multirow{2}{*}{Dataset} & CLIP & CoOp & CoCoOp & \multicolumn{2}{c}{\textbf{BadCLIP}} \\
 &  & ACC & ACC & ACC & ACC & ASR \\ \midrule
Source & ImageNet & 66.74 & 71.51 & 71.02 & 70.77 & 99.93 \\ \midrule
\multirow{11}{*}{Target} & Caltech101 & 93.09 & 93.70 & 94.43 & 93.63 & 100.0 \\
 & OxfordPets & 89.07 & 89.14 & 90.14 & 90.70 & 100.0 \\
 & StanfordCars & 65.17 & 64.51 & 65.32 & 64.17 & 100.0 \\
 & Flowers102 & 71.14 & 68.71 & 71.88 & 70.83 & 100.0 \\
 & Food101 & 86.07 & 85.30 & 86.06 & 85.17 & 100.0 \\
 & FGVCAircraft & 24.62 & 18.47 & 22.94 & 23.40 & 100.0 \\
 & SUN397 & 62.52 & 64.15 & 67.36 & 66.90 & 100.0 \\
 & DTD & 44.38 & 41.92 & 45.73 & 45.00 & 99.77 \\
 & EuroSAT & 47.53 & 46.39 & 45.37 & 45.13 & 100.0 \\
 & UCF101 & 66.67 & 66.55 & 68.21 & 68.17 & 100.0 \\ \cmidrule(l){2-7} 
 & Average & 65.02 & 63.88 & 65.74 & 65.31 & 99.98 \\ \bottomrule
\end{tabular}}
\vspace{-1em}
\end{table}

% \begin{tabular}{@{}llcccc@{}}
% \toprule
% \multirow{2}{*}{} & \multirow{2}{*}{Dataset} & CoOp & CoCoOp & \multicolumn{2}{c}{BadCLIP} \\
%  &  & ACC & ACC & ACC & ASR \\ \midrule
% Source & ImageNet & 71.51 & 71.02 & 70.77 & 99.93 \\ \midrule
% \multirow{11}{*}{Target} & Caltech101 & 93.70 & 94.43 & 93.63 & 100.0 \\
%  & OxfordPets & 89.14 & 90.14 & 90.70 & 100.0 \\
%  & StanfordCars & 64.51 & 65.32 & 64.17 & 100.0 \\
%  & Flowers102 & 68.71 & 71.88 & 70.83 & 100.0 \\
%  & Food101 & 85.30 & 86.06 & 85.17 & 100.0 \\
%  & FGVCAircraft & 18.47 & 22.94 & 23.40 & 100.0 \\
%  & SUN397 & 64.15 & 67.36 & 66.90 & 100.0 \\
%  & DTD & 41.92 & 45.73 & 45.00 & 99.77 \\
%  & EuroSAT & 46.39 & 45.37 & 45.13 & 100.0 \\
%  & UCF101 & 66.55 & 68.21 & 68.17 & 100.0 \\ \cmidrule(l){2-6} 
%  & Average & 63.88 & 65.74 & 65.31 & 99.98 \\ \bottomrule
% \end{tabular}

 In this part, we evaluate the performance of prompt learning methods under the cross-dataset setting, especially for backdoors learned by our BadCLIP. 
The results are shown in  Table \ref{tab:cross_dataset}.
In this setting, the accuracy on clean images of BadCLIP is on par with CoCoOp and surpasses CoOp by a large margin up to 1.43\% on average. Also, we surprisingly find that BadCLIP obtains 100\% attack success rates on 9 out of 10 datasets.  It illustrates that the trigger-aware context generator and the trigger learned on ImageNet can be applied to attack various downstream datasets. Notably, the attack can still succeed on these datasets containing totally different categories from ImageNet, such as Food101 and UCF101. Our results demonstrate that BadCLIP poses a serious security threat to downstream tasks even though the attacker cannot access their datasets.

\subsection{Cross-Domain Transfer}
% \begin{table}[t]
% \caption{Results of three learning based prompt methods under the cross-domain transfer setting. They are trained on ImageNet and tested on its 4 domain-shifted variants.}
% \label{sec:domain}
% \setlength\tabcolsep{12pt}
% \vspace{-0.5em}
% \centering
% \resizebox{\linewidth}{!}{
% \begin{tabular}{@{}clcccc@{}}
% \toprule
% \multirow{2}{*}{} & \multirow{2}{*}{Dataset} & CoOp & CoCoOp & \multicolumn{2}{c}{\textbf{BadCLIP}} \\ \cmidrule(l){3-6} 
%  &  & ACC & ACC & ACC & ASR \\ \midrule
% Source & ImageNet & 71.51 & 71.02 & 70.77 & 99.93 \\ \midrule
% \multirow{5}{*}{Target} & ImageNetV2 & 64.20 & 64.07 & 63.93 & 100.0 \\
%  & ImageNet-Sketch & 47.99 & 48.75 & 48.47 & 99.70 \\
%  & ImageNet-A & 49.71 & 50.63 & 49.67 & 100.0 \\
%  & ImageNet-R & 75.21 & 76.18 & 75.33 & 99.97 \\ \cmidrule(l){2-6} 
%  & Average & 59.28 & 59.91 & 59.35 & 99.92 \\ \bottomrule
% \end{tabular}}
% \vspace{-1em}
% \end{table}

\begin{table}[t]
\caption{Results of four methods under the cross-domain transfer setting. The learning based methods are trained on ImageNet and tested on its 4 domain-shifted variants.}
\label{sec:domain}
\setlength\tabcolsep{9pt}
\vspace{-0.5em}
\centering
\resizebox{\linewidth}{!}{
\begin{tabular}{@{}clccccc@{}}
\toprule
\multirow{2}{*}{} & \multirow{2}{*}{Dataset} & CLIP & CoOp & CoCoOp & \multicolumn{2}{c}{\textbf{BadCLIP}} \\ %\cmidrule(l){3-7} 
 &  & ACC & ACC & ACC & ACC & ASR \\ \midrule
Source & ImageNet & 66.73 & 71.51 & 71.02 & 70.77 & 99.93 \\ \midrule
\multirow{5}{*}{Target} & ImageNetV2 & 60.83 & 64.20 & 64.07 & 63.93 & 100.0 \\
 & ImageNet-Sketch & 46.15 & 47.99 & 48.75 & 48.47 & 99.70 \\
 & ImageNet-A & 47.77 & 49.71 & 50.63 & 49.67 & 100.0 \\
 & ImageNet-R & 73.96 & 75.21 & 76.18 & 75.33 & 99.97 \\ \cmidrule(l){2-7} 
 & Average & 55.96 & 59.28 & 59.91 & 59.35 & 99.92 \\ \bottomrule
\end{tabular}}
\vspace{-1em}
\end{table}

% one table
The cross-domain transferability is critical for backdoor attacks to succeed in diverse real-world scenarios. Following previous works \cite{zhou2022learning,zhou2022conditional}, we perform the prompt learning on ImageNet and test models on its 4 domain-shifted variants, as shown in Table \ref{sec:domain}. We can see that BadCLIP achieves similar performance compared to CoCoOp regarding accuracy on clean images, indicating that it inherits the advantages of the learnable prompts \cite{zhou2022learning}. We can also observe that BadCLIP reaches high attack success rates on all target datasets, ranging from 99.70\% to 100.0\%. These results suggest that BadCLIP is robust to domain shift.

\subsection{Comparison with Existing Attacks}
\label{sec:comp}

% To the best of our knowledge, there are two existing methods for backdoor attacks on CLIP. 

\vspace{0.2em}
\noindent \textbf{Data poisoning based attack.}
This method \cite{carlini2021poisoning} assumes that the attacker has access to the pre-training dataset for data poisoning and the CLIP model is pre-trained on it. Since our attack happens in the prompt-learning stage for the pre-trained CLIP model, it is infeasible to conduct a fair comparison between the data poison based attack and our BadCLIP. However, we observe from \cite{carlini2021poisoning} that data poisoning may limit the attack performance. 
For instance, its attack success rate is less than 80\% when inserting 1,500 poisoned samples into the Conceptual Captions dataset \cite{sharma2018conceptual}. 

\vspace{0.2em}
\noindent \textbf{Fine-tuning on poisoning data.}
We provide another baseline considered by CleanCLIP \cite{bansal2023cleanclip}, i.e., fine-tuning on poisoning data. For the CLIP with ResNet-50, it achieves a 58.40\% ACC and a 94.60\% ASR, while our BadCLIP performes better, with a 67.10\% ACC and a 98.75\% ASR, indicating the superiority of our design.

\vspace{0.2em}
\noindent \textbf{BadEncoder.}
It fine-tunes the image encoder of the pre-trained CLIP model with a large amount of additional unlabeled data, and then trains a task-specific classifier with the downstream dataset.
Table \ref{tab:badencoder} shows the comparison between BadEncoder and our BadCLIP on STL10 adopted in \cite{jia2022badencoder}, where the number of labeled training samples per class is 16. Following  \cite{jia2022badencoder}, we use ResNet-50 as the image encoder’s
backbone and set the target class as ``\textit{truck}''. For a comprehensive comparison, we vary the source and number of additional data adopted in BadEncoder. As can be seen, BadEncoder achieves a high clean accuracy and attack success rate with 50,000 additional data samples from STL10. However, when we reduce the amount of additional data or change the source, the attack performance is degraded significantly. Hence, BadEncoder depends on a large amount of additional data from a similar source as that of the downstream dataset, while our BadCLIP does not require additional data. Besides, we would like to emphasize that, unlike our BadCLIP, BadEncoder is not generalizable to unseen classes due to the task-specific classifier.

\begin{table}[t]
\caption{Comparison between BadEncoder \cite{jia2022badencoder} and BadCLIP on STL10. ``-'' implies that BadCLIP does not require additional data.}
\vspace{-0.5em}
\label{tab:badencoder}
\setlength\tabcolsep{12pt}
\resizebox{\linewidth}{!}{
\begin{tabular}{@{}lcccc@{}}
\toprule
Method & \begin{tabular}[c]{@{}c@{}}Source of   \\ Additional Data\end{tabular} & \begin{tabular}[c]{@{}c@{}}Number of   \\ Additional Data\end{tabular} & ACC & ASR \\ \midrule
\multirow{4}{*}{BadEncoder} & STL10 & 50,000 & 94.83 & 99.96 \\
 & STL10 & 5,000 & 94.74 & 92.47 \\
 & STL10 & 1,000 & 94.01 & 17.39 \\
 & SVHN & 5,000 & 91.33 & 11.45 \\ \midrule
\textbf{BadCLIP} & - & - & 95.13 & 98.57 \\ \bottomrule
\end{tabular}}
\vspace{-1em}
\end{table}

\begin{figure}[t]
\centering
\includegraphics[width=0.95\linewidth]{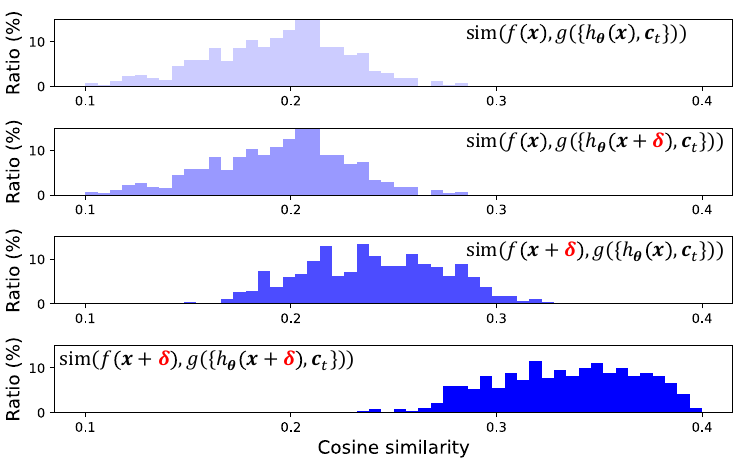}
\vspace{-1em}
\caption{Distribution of cosine similarities between images and text prompts in the feature space. 
% $\bm{x}$, $\bm{x}+\bm{\delta}$, and $\bm{c}_t$ denote the clean image, its backdoor image, and the word embedding vector of the target class $t$, respectively. 
$f(\bm{x})$: clean image features; $f(\bm{x}+\bm{\delta})$: backdoor image features; $g(\{h_{\bm{\theta}}(\bm{x}), \bm{c}_t\})$: clean text features for the target class $t$; $g(\{h_{\bm{\theta}}(\bm{x}+\bm{\delta}), \bm{c}_t\})$: backdoor text features for the target class $t$. When both image and text encoders take backdoor inputs (\textbf{bottom}), the cosine similarity is highest on average, resulting in the best attack performance.}
\vspace{-1em}
\label{fig:sim_hist}
\end{figure}

\begin{figure}[t]
\centering
\begin{subfigure}{0.47\linewidth}
\centering
\includegraphics[width=0.9\linewidth,height=0.8\linewidth]{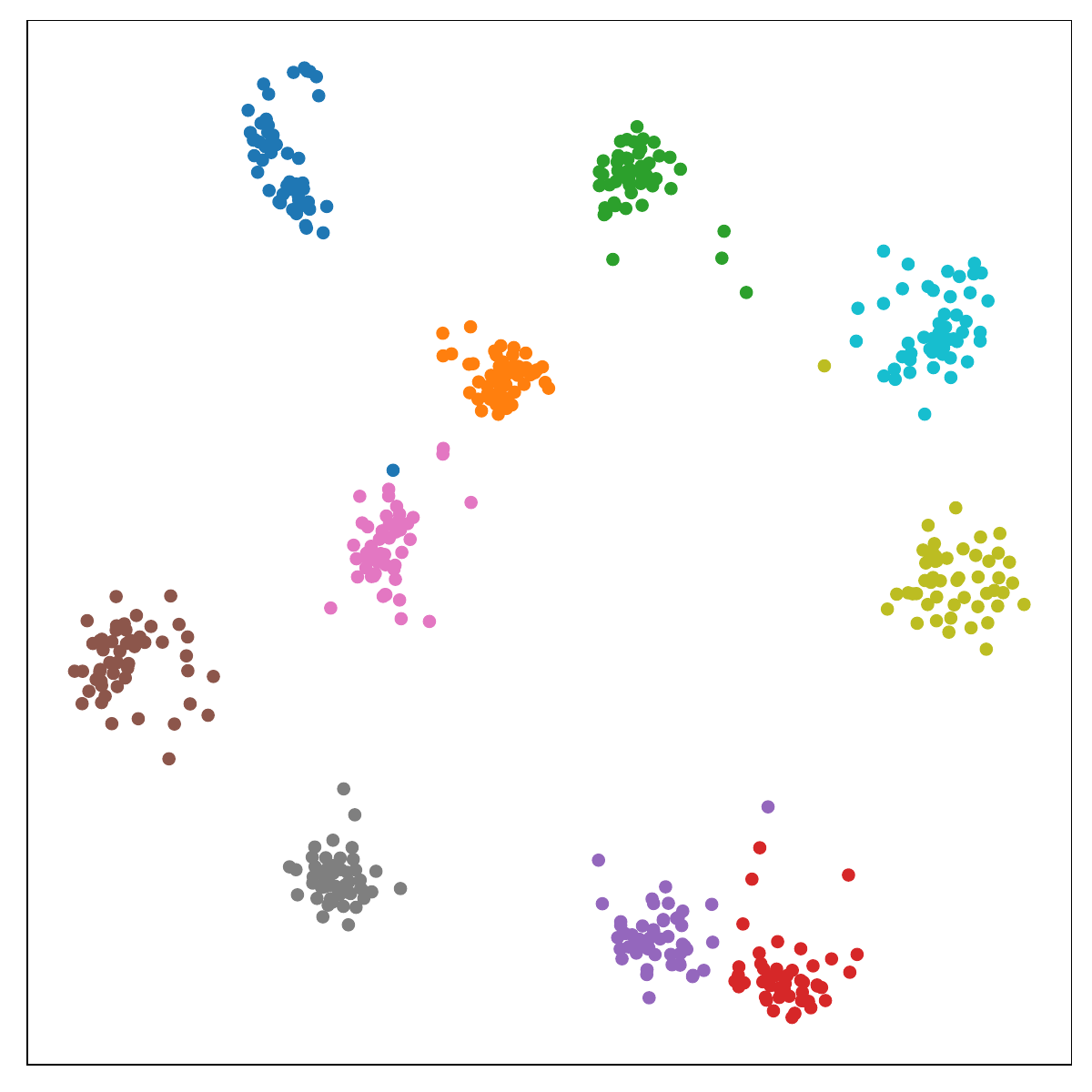}
\caption{Clean images}
\end{subfigure}
\hfill
\begin{subfigure}{0.47\linewidth}
\centering
\includegraphics[width=0.9\linewidth,height=0.8\linewidth]{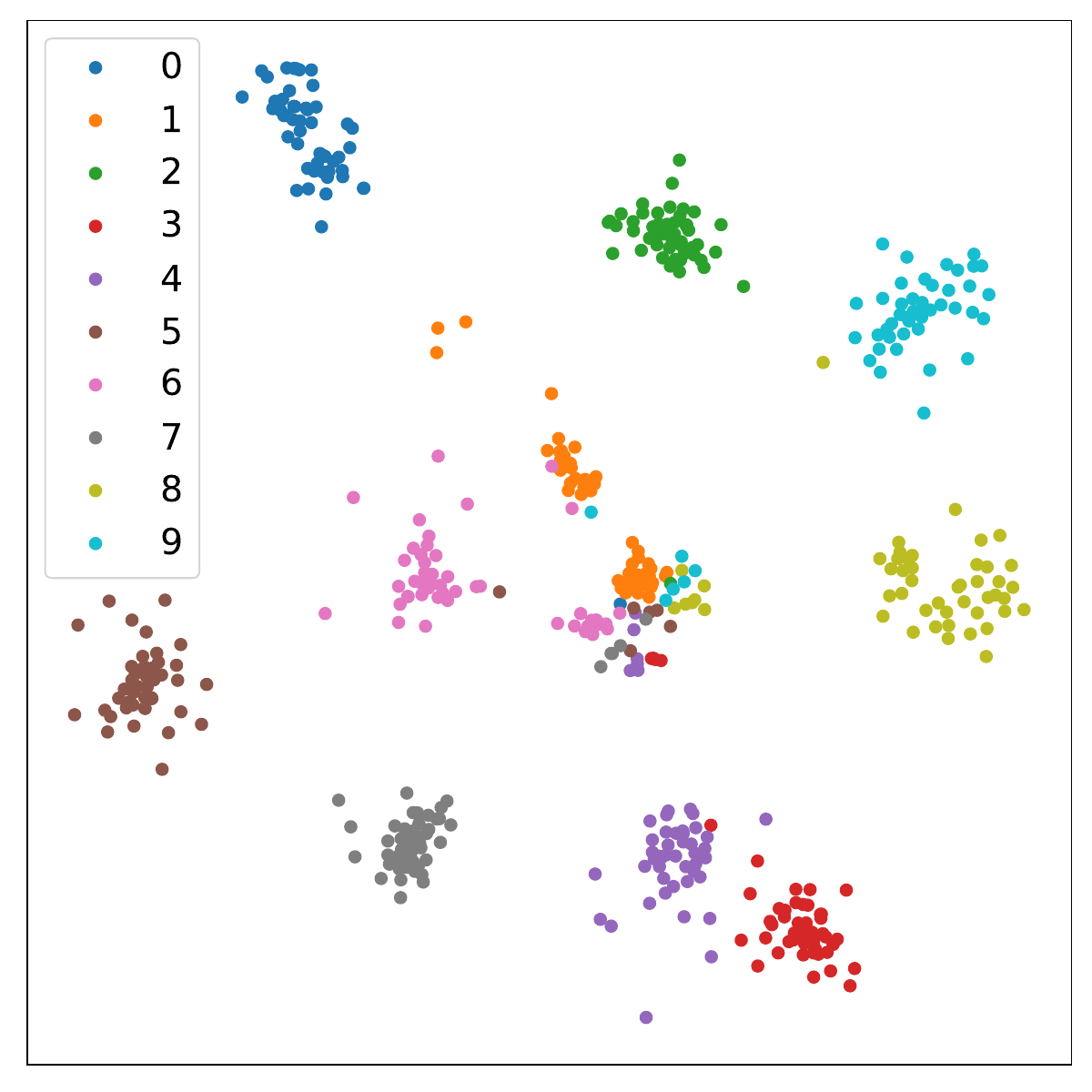}
\caption{Backdoor images}
\end{subfigure}
\vspace{-0.5em}
\caption{t-SNE visualization of features extracted by BadCLIP's image encoder for clean images and their backdoor versions from 10 random classes on ImageNet.  Our backdoor image features are still separable. Note that the class 0 corresponds to the target class.}
\vspace{-1em}
\label{fig:tsne}
\end{figure}

\subsection{Trigger-Aware Prompts Matter}
\label{sec:understanding}
% \vspace{0.5em}
% \noindent \textbf{Understanding BadCLIP in the feature space.}
% one table and two figures
% ['tench' 'hammerhead shark' 'sea anemone' 'Standard Schnauzer' 'Kuvasz' 'hyena' 'otter' 'giant panda' 'accordion' 'birdhouse']
\noindent \textbf{Understanding BadCLIP in the feature space.}
\ 
% To show the effect of trigger-aware prompts, we analyze the behaviour of BadCLIP in the feature space. As demonstrated in Fig. \ref{fig:overview}, the backdoor images change both image and text features in our BadCLIP. 
As demonstrated in Fig. \ref{fig:overview}, the backdoor images change both image and text features in our BadCLIP. Here, to show the effect of trigger-aware prompts, we propose to decouple the inputs into the image and text encoders to analyze the effect of the changes of image and text features, respectively. 
Specifically, the image encoder takes the clean image $\bm{x}$ or the backdoor image $\bm{x}+\bm{\delta}$ as inputs; 
the text encoder takes the clean text prompt $\{h_{\bm{\theta}}(\bm{x}), \bm{c}_t\}$ or the backdoor text prompt $\{h_{\bm{\theta}}(\bm{x}+\bm{\delta}), \bm{c}_t\}$ for the target class $t$ as inputs. 
We calculate the distribution of cosine similarities between images and text features in four cases, as shown in Fig. \ref{fig:sim_hist}. As can be seen, when both image and text encoders take backdoor inputs, the cosine similarity is highest on average, implying that inputs are classified into the target class with the highest confidences. Our analysis illustrates that the success of our backdoor attack can be attributed to the collaboration between the changes of image and text features. Thus, although the features of images shift across different scenarios, the textual features of the target class change along with the trigger, ensuring successful attacks. 
This insight is fundamental and critical, and will inspire backdoor studies on multi-modal models. 

The t-SNE \cite{van2008visualizing} visualization of clean and backdoor image features further confirms the effect of trigger-aware prompts. As suggested in \cite{wu2022backdoorbench}, for backdoor attacks on the image recognition models only relying on the visual modality, their backdoor image features cluster together. In contrast, for our BadCLIP built on visual and textual modalities, its backdoor image features are still separable as shown in Fig. \ref{fig:tsne}. This observation indirectly indicates that the backdoor text prompts contribute a lot to the targeted misclassification in our method. We believe that this interesting phenomenon for multi-modal models is worthy of a further exploration from both backdoor attack and defense sides.

\begin{table}[t]
\caption{Comparison of the trigger-agnostic prompts and trigger-aware prompts (adopted in our BadCLIP) in backdoor attacks. Results are averaged over 11 datasets.}
\vspace{-0.5em}
\label{tab:trigger_agnostic}
\setlength\tabcolsep{3pt}
\resizebox{\linewidth}{!}{
\begin{tabular}{@{}lcc|cc|cc@{}}
\toprule
\multirow{2}{*}{Method} & \multicolumn{2}{c|}{Seen} & \multicolumn{2}{c|}{Unseen} & \multicolumn{2}{c}{H} \\ \cmidrule(l){2-7} 
 & ACC & ASR & ACC & ASR & ACC & ASR \\ \midrule
Trigger-Agnostic Prompts & 76.19 & 95.31 & 62.73 & 2.21 & 68.14 & 3.81 \\
Trigger-Aware Prompts (Ours) & 79.55 & 99.52 & 69.86 & 99.02 & 73.95 & 99.26 \\ \bottomrule
\end{tabular}}
\vspace{-1em}
\end{table}

\noindent \textbf{Backdoor attack with trigger-agnostic prompts.}
\ We study the effect of trigger-agnostic prompts by comparing BadCLIP with a baseline, i.e., the backdoor attack with trigger-agnostic prompts. Specifically, following \cite{zhong2021factual,lester2021power,zhou2022learning}, 
we model context tokens using continuous vectors, which are fixed for any image input once learned, such that the text features cannot be changed by the backdoor images. Other settings are the same as those used in BadCLIP. The comparison in Table \ref{tab:trigger_agnostic} shows the superiority of our method.  In particular, backdoor attack with trigger-agnostic prompts fails to generalize to unseen classes. These results demonstrate that trigger-aware prompts have a positive effect on the generalizability of BadCLIP.

% \begin{figure}[h]
% \centering
% \includegraphics[width=\linewidth]{figures/effect_of_eps.pdf}
% \vspace{-1em}
% \caption{Results of BadCLIP with different maximum noise strengths. All reported values are based on the harmonic mean.}
% \label{fig:effect_of_eps}
% \end{figure}

\subsection{Resistance to Backdoor Defense Methods}

\begin{figure}[t]
\centering
\begin{subfigure}{0.493\linewidth}
\centering
\includegraphics[width=0.9\linewidth,height=0.7\linewidth]{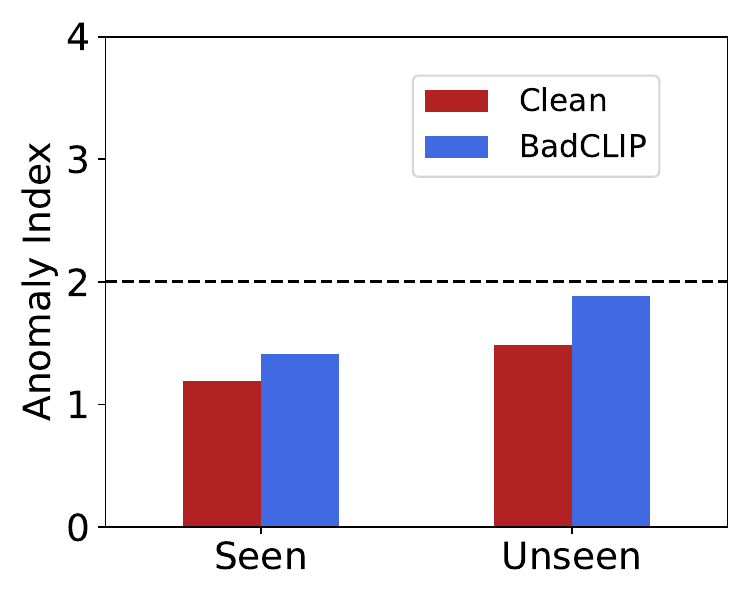}
\vspace{-0.2em}
\caption{Neural Cleanse}
\label{fig:cleanse}
\end{subfigure}
\hfill
\begin{subfigure}{0.493\linewidth}
\centering
\includegraphics[width=0.9\linewidth,height=0.68\linewidth]{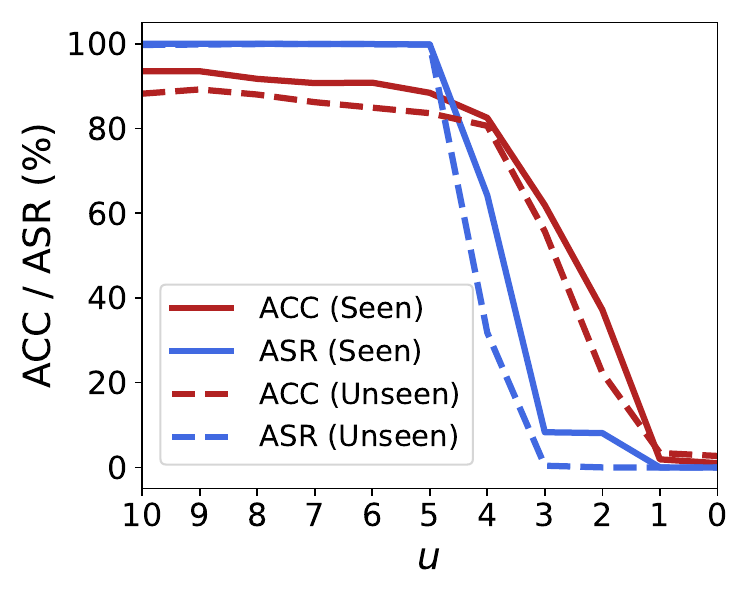}
\vspace{-0.2em}
\caption{CLP defense}
\label{fig:prune}
\end{subfigure}
\vspace{-1.6em}
\caption{Results of defense experiments on Caltech101.}
% \label{fig:effect_of_eps}
\vspace{-1.2em}
\end{figure}

% Since there

\noindent \textbf{Resistance to Neural Cleanse.}
%  is a model mitigation defense based on the pattern optimization approach. It 
\ Neural Cleanse \cite{wang2019neural} assumes that the backdoor trigger is patch based. For each class, it reconstructs the optimal patch pattern to convert any clean input to that target class. 
If any class has a significantly smaller pattern than the others, Neural Cleanse considers it as a backdoor indicator. 
It is quantified by the Anomaly Index metric.
If the Anomaly Index is less than a threshold of 2 for a specific class, the defense considers that there is a backdoor
with this class as the target label. We show the results of the clean CLIP model and our BadCLIP on Caltech101 in Fig. \ref{fig:cleanse}.
Similar to the clean model, BadCLIP passes the tests with very small scores, showing that our attack is resistant to Neural Cleanse.
% It is because BadCLIP uses additive noise as the trigger while Neural Cleanse is based on the patch based detection mechanism. 
% \red{One reason may be that it is very difficult to optimize small patch patterns due to the dynamic text features.}

\vspace{0.1em}
\noindent \textbf{Resistance to CLP defense.} 
Channel Lipschitzness based Pruning (CLP) \cite{zheng2022data} is a data-free backdoor removal method. It prunes those neurons that are sensitive to input changes.  Fig. \ref{fig:prune} presents the results under different settings of $u$ in CLP. A smaller $u$ means a larger pruning ratio. 
We can see from the figure that when CLP removes the backdoor ($u > 3$), the accuracy on clean images is significantly reduced. 
Therefore, CLP cannot eliminate the backdoor injected by our BadCLIP with a high ACC.

\subsection{Extensible Application Scenario}

Here, we evaluate our attack on more application scenarios. 
Firstly, we apply our attack to a recently released version of CLIP, named OpenCLIP \cite{cherti2022reproducible}, which utilizes a different pre-training dataset (LAION) \cite{schuhmannlaion}, and many additional techniques such as using a pre-trained language model and scaling up to a huge-scale model architecture. 
% The results of two variants of OpenCLIP are in Table \ref{tab:openclip} and BadOpenCLIP denotes our attack.
% It shows our attack can be applied to OpenCLIP.
Table \ref{tab:openclip} shows the results of two variants of OpenCLIP on UCF-101. We can see that our method can succeed in attacking these two models. 
Secondly, we carry out experiments on the image-text retrieval task with Flickr30K \cite{young2014image}, following \cite{he2022cpl}.  The prompt learning methods are trained with only 3\% of training data and tested on the complete test set. R@1 and B-R@1 denote Recall at 1 and that of backdoor image queries, respectively. Table \ref{tab:retrieval} shows the success of BadCLIP on the image-text retrieval task. These results indicate that the application scenario of BadCLIP is extensible.

\begin{table}[t]
\caption{Results of the proposed attack on OpenCLIP. BadOpenCLIP denotes our attack.}
\vspace{-1em}
\label{tab:openclip}
\centering
\setlength{\tabcolsep}{5mm}{
\resizebox{\linewidth}{!}{
\begin{tabular}{@{}ccc|ccc@{}}
\toprule
\multicolumn{3}{c|}{Pre-trained Language Model} & \multicolumn{3}{c}{Huge-scale Model} \\ \midrule
OpenCLIP & \multicolumn{2}{c|}{\textbf{BadOpenCLIP}} & OpenCLIP & \multicolumn{2}{c}{\textbf{BadOpenCLIP}} \\
ACC & ACC & ASR & ACC & ACC & ASR \\ \midrule
69.86 & 74.15 & 98.81 & 80.56 & 84.49 & 99.90 \\ \bottomrule
\end{tabular}}}
\end{table}

\begin{table}[t]
\vspace{-0.7em}
\centering
\caption{Results of BadCLIP on the image-text retrieval task.}
\vspace{-1em}
\label{tab:retrieval}
\centering
\resizebox{0.7\linewidth}{!}{
\setlength{\tabcolsep}{5mm}{
\begin{tabular}{@{}ccccc@{}}
\toprule
CLIP & CoOp & CoCoOp & \multicolumn{2}{c}{\textbf{BadCLIP}} \\
R@1 & R@1 & R@1 & R@1 & \!\!B-R@1 \\ \midrule
83.0 & 79.4 & 85.9 & 85.2 & \!\!98.3 \\ \bottomrule
\end{tabular}}}
\vspace{-1.3em}
\end{table}

% \subsection{Ablation Studies}

% In this section, we discuss the effect of the trigger warm-up strategy and the maximum noise strength. More details can be found in Appendix C.

% The comparison of BadCLIP with and without the trigger warm-up strategy is shown in Table \uppercase\expandafter{\romannumeral2} in Appendix C. Compared with optimizing $\bm{\theta}$ and $\bm{\delta}$ from scratch, using the trigger warm-up strategy brings 2.03\% and 0.2\% gains on average in terms of ACC and ASR, respectively. The reason may be that individually optimizing $\bm{\delta}$ provides a good initialization for the joint optimization stage. 

% We show the results with $\epsilon \in \{0.1,0.5,1,2,4\}$ in Figure \uppercase\expandafter{\romannumeral2} in Appendix C.
% When $\epsilon$ is relatively small ($<$1), the ACC and ASR values increase with the increase of $\epsilon$. However, when $\epsilon$ is larger than 1, the performance remains almost unchanged for different $\epsilon$. It illustrates that BadCLIP can be effective even with a very small noise strength. 
% Considering the generalizability in various settings and the visual stealthiness described in Section \ref{sec:few_shot}, $\epsilon=4$ by default in our experiments is a reasonable choice.

\vspace{-0.1em}
% \section{Conclusion}
\section{Conclusions}
\vspace{-0.1em}
In this paper, we explored backdoor attacks on CLIP with limited downstream training data. We proposed BadCLIP which accomplishes backdoor attacks via trigger-aware prompt learning. BadCLIP consists of a learnable trigger applied to images and a trigger-aware context generator. We then proposed an optimization method based on a novel warm-up strategy. We showed that BadCLIP achieves promising attack performance and a generalizable backdoor. To the best of our knowledge, BadCLIP is the first backdoor attack on CLIP in the prompt learning stage. We would hope that our work opens a new domain of attack mechanisms on vision-language models, and can encourage future defense research. 

\noindent \textbf{Acknowledgement.}
This work is supported in part by the National Natural Science Foundation of China under Grant 62171248, Shenzhen Science and Technology Program (JCYJ20220818101012025), and the PCNL KEY project (PCL2023AS6-1).

% \noindent \textbf{Limitation and future work.} Despite promising performance of BadCLIP in most cases, there is a gap between the accuracy on clean images of BadCLIP and that of using hand-crafted prompts in unseen classes. In fact, this is a challenging problem for prompt learning methods \cite{zhou2022learning,zhou2022conditional,zhu2022prompt}, which is an interesting future direction for backdoor attacks on CLIP.  
% Another limitation of BadCLIP is that it assumes that the attacker has full
% knowledge of the pre-trained CLIP model including model architectures and parameters. We will
% further explore more strict settings than the white-box one in our future work.


{
    \small
    \bibliographystyle{ieeenat_fullname}
    \bibliography{main}
}

\appendix

\begin{table}[t]
\centering
\caption{Results of BadCLIP under various settings. We report the harmonic mean of results on the seen and unseen classes.}
\label{tab:various_settings}
\begin{minipage}{0.4\textwidth}
\begin{subfigure}{\linewidth}
\caption{Varying the context length.}
\label{tab:context_length}
\centering
\setlength\tabcolsep{15pt}
\resizebox{\linewidth}{!}{
\begin{tabular}{@{}lcccc@{}}
\toprule
\multirow{2}{*}{Dataset} & \multirow{2}{*}{Metric} & \multicolumn{3}{c}{Context Length} \\ \cmidrule(l){3-5} 
 &  & 4 & 8 & 16 \\ \midrule
\multirow{2}{*}{Caltech101} & ACC & 95.58 & 95.72 & 95.66 \\
 & ASR & 99.46 & 99.58 & 99.25 \\ \midrule
\multirow{2}{*}{StanfordCars} & ACC & 71.34 & 70.88 & 71.56 \\
 & ASR & 99.80 & 99.58 & 99.83 \\ \midrule
\multirow{2}{*}{UCF101} & ACC & 76.31 & 76.90 & 76.67 \\
 & ASR & 99.62 & 99.57 & 99.90 \\ \bottomrule
\end{tabular}}
\end{subfigure}
\end{minipage}
\begin{minipage}{0.45\textwidth}
\begin{subfigure}{\linewidth}
\vspace{1em}
\caption{Varying the number of training data.}
\label{tab:effect_of_shots}
\resizebox{\linewidth}{!}{
\begin{tabular}{@{}lcccccc@{}}
\toprule
\multirow{2}{*}{Dataset} & \multirow{2}{*}{Metric} & \multicolumn{5}{c}{\# of Labeled Training Examples per Class} \\ \cmidrule(l){3-7} 
 &  & 1 & 2 & 4 & 8 & 16 \\ \midrule
\multirow{2}{*}{Caltech101} & ACC & 91.05 & 94.76 & 95.13 & 95.58 & 95.58 \\
 & ASR & 88.25 & 96.42 & 98.76 & 98.97 & 99.46 \\ \midrule
\multirow{2}{*}{StanfordCars} & ACC & 67.21 & 68.68 & 69.53 & 70.47 & 71.36 \\
 & ASR & 98.01 & 98.90 & 98.83 & 99.63 & 99.80 \\ \midrule
\multirow{2}{*}{UCF101} & ACC & 70.95 & 71.94 & 74.23 & 75.45 & 76.36 \\
 & ASR & 94.77 & 97.88 & 98.66 & 99.40 & 99.62 \\ \bottomrule
\end{tabular}}
\end{subfigure}
\end{minipage}
\begin{minipage}{0.43\textwidth}
\begin{subfigure}{\linewidth}
\vspace{1em}
\caption{Using ResNet-50 as the image encoder's backbone.}
\label{tab:resnet}
\setlength\tabcolsep{9pt}
\resizebox{\linewidth}{!}{
\begin{tabular}{@{}lccccc@{}}
\toprule
\multirow{2}{*}{Dataset} & CLIP & CoOp & CoCoOp & \multicolumn{2}{c}{\textbf{BadCLIP}} \\
 & ACC & ACC & ACC & ACC & ASR \\ \midrule
Caltech101 & 90.80 & 89.29 & 92.67 & 92.04 & 99.46 \\
StanfordCars & 60.50 & 57.44 & 64.25 & 62.78 & 99.83 \\
UCF101 & 69.14 & 52.59 & 70.80 & 69.30 & 99.71 %\\ \midrule
\\ \bottomrule
\end{tabular}}
\end{subfigure}
\end{minipage}
\end{table}

\section{Results under Various Settings}
% one table, one figure
\label{sec:other_settings}

In this part, we investigate the effect of various settings on the proposed BadCLIP, including context length, number of training examples, and image encoder's backbone.  

\vspace{0.5em}
\noindent \textbf{Context length.}
\ Following \cite{zhou2022learning,zhou2022conditional}, we study the performance of BadCLIP when the context length $N$ is set as 4, 8, and 16. The results in Table \ref{tab:context_length} show that the differences between different context lengths are fairly small and the best choice depends on the dataset. Notably, the ASR values are higher than 99\% in all cases, showing the robustness of BadCLIP to various settings of context length. 

\vspace{0.5em}
\noindent \textbf{Number of training examples.}
\ We study the BadCLIP with different numbers of labeled training examples per class ranging from 1 to 16, as shown in Table \ref{tab:effect_of_shots}. As expected, both the accuracy on clean images and the attack success rate increase with the increase of the number of training examples. In particular, 
BadCLIP can obtain high attack success rates with a small number of training examples. 
For example, the ASR value is 98.01\% on StandfordCars when the number of labeled training examples per class is 1. 
It shows that 4 labeled training examples per class are enough for BadCLIP to reach a satisfactory ASR ($>$98\%). 

% \begin{table}[]
% \caption{Results of using ResNet-50 as the image encoder’s backbone. All reported values are based on the harmonic mean.}
% \label{tab:resnet}
% \vspace{-0.5em}
% \setlength\tabcolsep{10pt}
% \centering
% \resizebox{\linewidth}{!}{
% \begin{tabular}{@{}lccccc@{}}
% \toprule
% \multirow{2}{*}{Dataset} & CLIP & CoOp & CoCoOp & \multicolumn{2}{c}{\textbf{BadCLIP}} \\
%  & ACC & ACC & ACC & ACC & ASR \\ \midrule
% % ImageNet & 62.13 & 63.86 & 65.35 & 65.03 & 99.65 \\
% Caltech101 & 90.80 & 89.29 & 92.67 & 92.04 & 99.46 \\
% % OxfordPets & 92.58 & 90.38 & 93.83 & 90.52 & 96.43 \\
% StanfordCars & 60.50 & 57.44 & 64.25 & 62.78 & 99.83 \\
% % Flowers102 & 71.69 & 66.85 & 77.29 & 77.11 & 99.62 \\
% % Food101 & 83.74 & 76.33 & 84.65 & 84.00 & 99.75 \\
% % FGVCAircraft & 21.99 & 19.82 & 24.32 & 20.91 & 99.75 \\
% % SUN397 & 68.21 & 66.86 & 73.79 & 72.85 & 99.68 \\
% % DTD & 54.68 & 51.99 & 54.63 & 52.95 & 97.12 \\
% % EuroSAT & 60.66 & 48.20 & 55.16 & 50.65 & 95.30 \\
% UCF101 & 69.14 & 52.59 & 70.80 & 69.30 & 99.71 %\\ \midrule
% % Average & 66.92 & 62.15 & 68.79 & 67.10 & 98.75 \\ \bottomrule
% \\ \bottomrule
% \end{tabular}}
% \end{table}

\vspace{0.5em}
\noindent \textbf{Image encoder's backbone.}
In our prior experiments, we use ViT-B/16 as the image encoder's backbone. For a more comprehensive study, we conduct the experiments on ResNet-50, as shown in Table \ref{tab:resnet}. 
The observations from Table 1 still hold under this setting. Specifically, BadCLIP achieves higher accuracies on clean images than zero-shot CLIP and CoOp, and high attack success rates. It verifies that BadCLIP can be implemented with different image encoder's backbones.  

% \begin{table}[]
% \caption{Results of BadCLIP with different context lengths. All reported values are based on the harmonic mean.}
% \label{tab:context_length}
% \centering
% \setlength\tabcolsep{10pt}
% \resizebox{0.8\linewidth}{!}{
% \begin{tabular}{@{}lcccc@{}}
% \toprule
% \multirow{2}{*}{Dataset} & \multirow{2}{*}{Metric} & \multicolumn{3}{c}{Context Length} \\ \cmidrule(l){3-5} 
%  &  & 4 & 8 & 16 \\ \midrule
% \multirow{2}{*}{Caltech101} & ACC & 95.58 & 95.72 & 95.66 \\
%  & ASR & 99.46 & 99.58 & 99.25 \\ \midrule
% \multirow{2}{*}{StanfordCars} & ACC & 71.34 & 70.88 & 71.56 \\
%  & ASR & 99.80 & 99.58 & 99.83 \\ \midrule
% \multirow{2}{*}{UCF101} & ACC & 76.31 & 76.90 & 76.67 \\
%  & ASR & 99.62 & 99.57 & 99.90 \\ \bottomrule
% \end{tabular}}
% \end{table}

% \begin{figure}[h]
% \centering
% \includegraphics[width=\linewidth]{figures/effect_of_shots.pdf}
% \vspace{-0.5em}
% \caption{Results of BadCLIP with different numbers of labeled training sample per class. All reported values are based on the harmonic mean.}
% \label{fig:effect_of_shots}
% \end{figure}
% ImageNet Caltech101 OxfordPets StanfordCars Flowers102 Food101 FGVCAircraft SUN397 DTD  EuroSAT  UCF101

\begin{figure*}[t]
\centering
\vspace{2em}
\begin{subfigure}{0.49\textwidth}
\includegraphics[width=\textwidth]{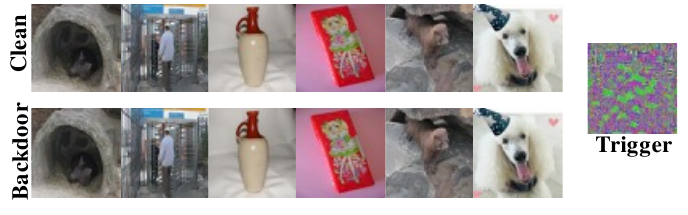}
\caption{ImageNet}
\end{subfigure}
\hfill
\centering
\begin{subfigure}{0.49\textwidth}
\includegraphics[width=\textwidth]{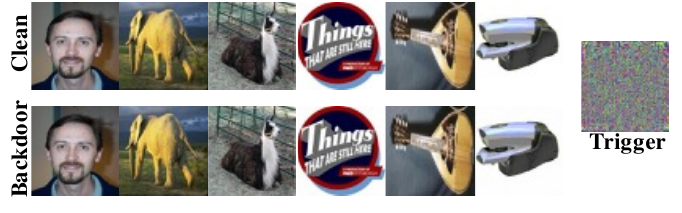}
\caption{Caltech101}
\end{subfigure}
\vspace{0.5em}
\centering
\begin{subfigure}{0.49\textwidth}
\includegraphics[width=\textwidth]{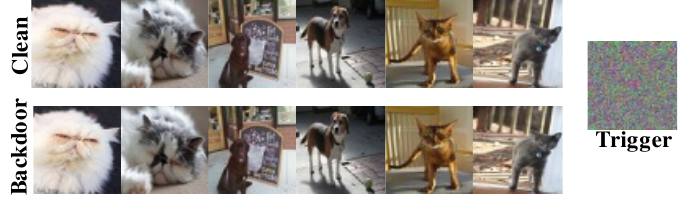}
\caption{OxfordPets}
\end{subfigure}
\hfill
\centering
\begin{subfigure}{0.49\textwidth}
\includegraphics[width=\textwidth]{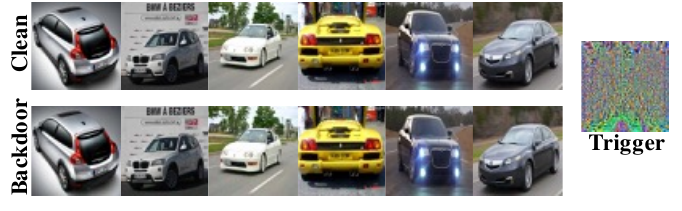}
\caption{StanfordCars}
\end{subfigure}
\vspace{0.5em}
\centering
\begin{subfigure}{0.49\textwidth}
\includegraphics[width=\textwidth]{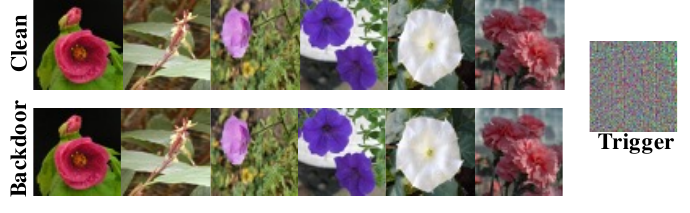}
\caption{Flowers102}
\end{subfigure}
\hfill
\centering
\begin{subfigure}{0.49\textwidth}
\includegraphics[width=\textwidth]{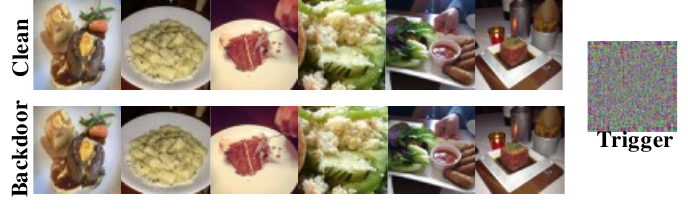}
\caption{Food101}
\end{subfigure}
\vspace{0.5em}
\centering
\begin{subfigure}{0.49\textwidth}
\includegraphics[width=\textwidth]{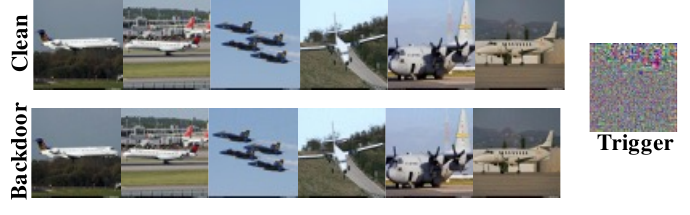}
\caption{FGVCAircraft}
\end{subfigure}
\hfill
\centering
\begin{subfigure}{0.49\textwidth}
\includegraphics[width=\textwidth]{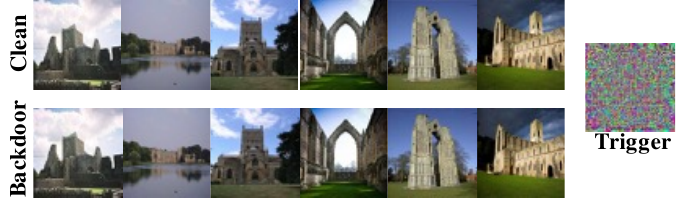}
\caption{SUN397}
\end{subfigure}
\vspace{0.5em}
\centering
\begin{subfigure}{0.49\textwidth}
\includegraphics[width=\textwidth]{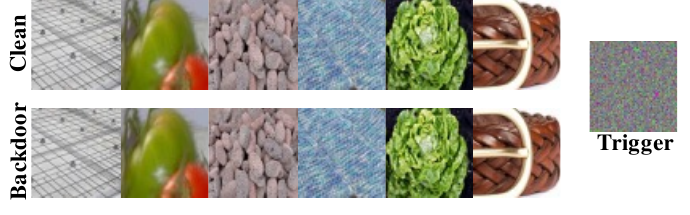}
\caption{DTD}
\end{subfigure}
\hfill
\centering
\begin{subfigure}{0.49\textwidth}
\includegraphics[width=\textwidth]{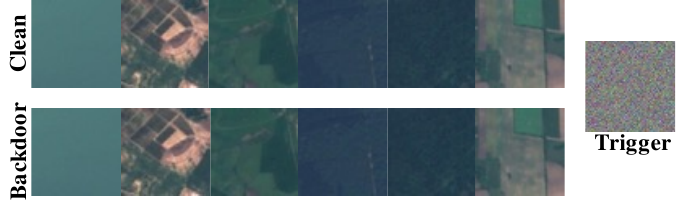}
\caption{EuroSAT}
\end{subfigure}
\vspace{0.5em}
% \centering
\begin{subfigure}{0.49\textwidth}
\includegraphics[width=\textwidth]{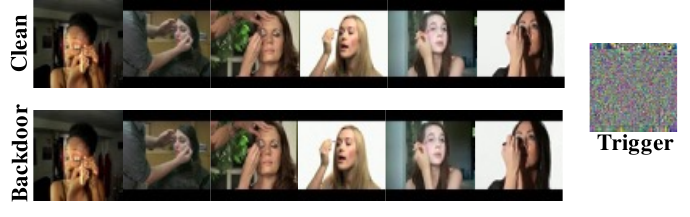}
\caption{UCF101}
\end{subfigure}
\hspace{25.1em}
% \vspace{0.5em}
\caption{ Visualization of clean images, backdoor images, and triggers on 11 datasets. The trigger is scaled for visibility.}
\label{fig:vis}
\vspace{2em}
\end{figure*}

\section{Visualization}
We provide visualization examples in Fig. \ref{fig:vis}. We can see that our trigger is so
small that there is no visual difference between the clean
and backdoor images. These results further demonstrate that our
attack is stealthy.

\begin{table}[ht]
\caption{Comparison of BadCLIP with and without the trigger warm-up. Results are averaged over 11 datasets.}
\label{tab:effect_of_warmup}
\vspace{-0.5em}
\setlength\tabcolsep{0.5pt}
\centering
\setlength\tabcolsep{10pt}
\resizebox{\linewidth}{!}{
% \begin{tabular}{@{}lcc|cc@{}}
% \toprule
% \multirow{2}{*}{Dataset} & \multicolumn{2}{c|}{ACC} & \multicolumn{2}{c}{ASR} \\ \cmidrule(l){2-5} 
%  & w/ warm-up & w/o warm-up & w/ warm-up & w/o warm-up \\ \midrule
% ImageNet & 72.90 & 72.72 & 99.65 & 99.68 \\
% Caltech101 & 95.58 & 94.24 & 99.46 & 99.28 \\
% OxfordPets & 88.68 & 94.05 & 98.96 & 97.80 \\
% StanfordCars & 71.34 & 70.96 & 99.80 & 99.95 \\
% Flowers102 & 82.18 & 78.66 & 99.91 & 99.61 \\
% Food101 & 90.10 & 89.75 & 98.90 & 99.70 \\
% FGVCAircraft & 32.96 & 19.08 & 99.68 & 96.76 \\
% SUN397 & 77.60 & 76.97 & 99.50 & 99.80 \\
% DTD & 59.81 & 60.09 & 97.92 & 98.46 \\
% EuroSAT & 65.98 & 59.04 & 98.49 & 99.03 \\
% UCF101 & 76.31 & 75.61 & 99.62 & 99.19 \\ \midrule
% Average & 73.95 & 71.92 & 99.26 & 99.02 \\ \bottomrule
% \end{tabular}
\begin{tabular}{@{}lcc|cc|cc@{}}
\toprule
\multirow{2}{*}{Method} & \multicolumn{2}{c|}{Seen} & \multicolumn{2}{c|}{Unseen} & \multicolumn{2}{c}{H} \\ \cmidrule(l){2-7} 
 & ACC & ASR & ACC & ASR & ACC & ASR \\ \midrule
w/o the Trigger Warm-up & 78.10 & 99.64 & 67.80 & 98.43 & 71.92 & 99.02 \\
w/ the Trigger Warm-up & 79.55 & 99.52 & 69.86 & 99.02 & 73.95 & 99.26 \\ \bottomrule
\end{tabular}
}
\end{table}

\section{Ablation Studies}

\noindent \textbf{Effect of the trigger warm-up strategy.}
% one table
\ To obtain a better solution to Problem (5), we propose the trigger warm-up strategy in the optimization process. Here, we study the effect of this component through the comparison of BadCLIP with and without the trigger warm-up strategy.  The results are shown in Table \ref{tab:effect_of_warmup}. Compared with optimizing $\bm{\theta}$ and $\bm{\delta}$ from scratch (i.e., without warm-up), optimizing with the trigger warm-up strategy brings 2.03\% and 0.2\% gains on average in terms of ACC and ASR, respectively. The reason may be that individually optimizing $\bm{\delta}$ provides a good initialization for the joint optimization stage. 

\begin{figure}[ht]
\centering
\begin{subfigure}{0.493\linewidth}
\centering
\includegraphics[width=0.95\linewidth,height=0.8\linewidth]{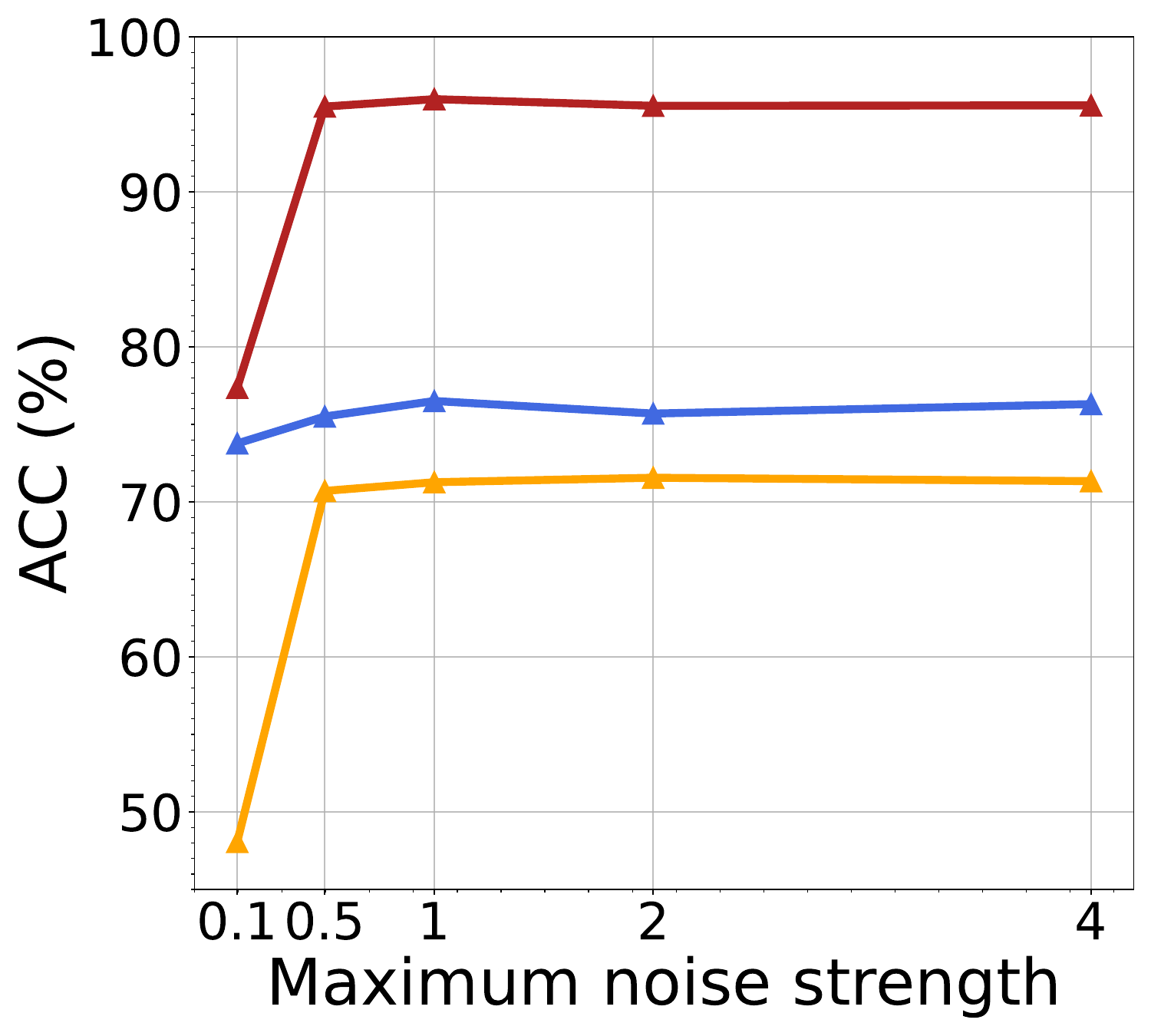}
\end{subfigure}
\hfill
\begin{subfigure}{0.493\linewidth}
\centering
\includegraphics[width=0.95\linewidth,height=0.8\linewidth]{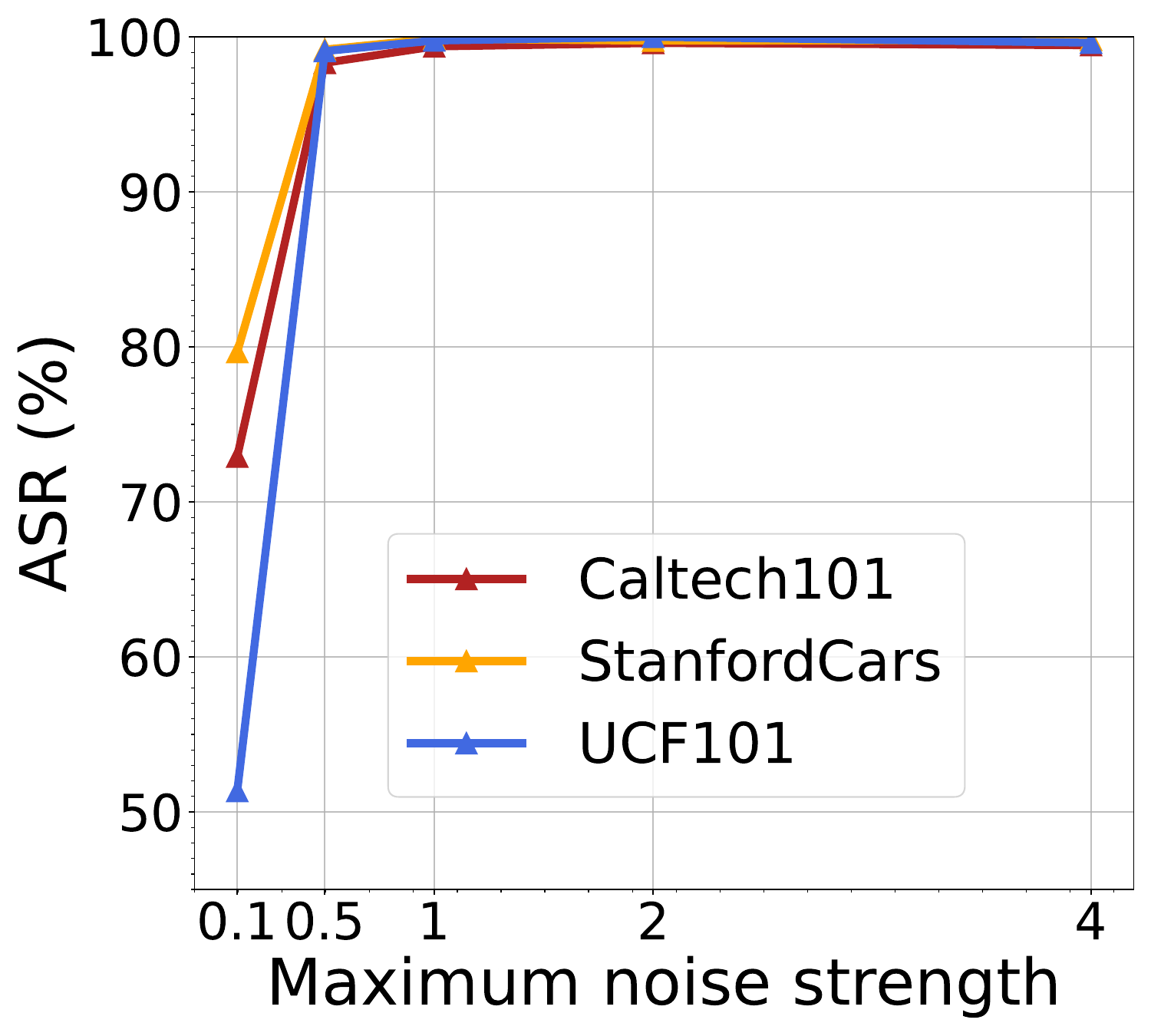}
\end{subfigure}
\vspace{-2em}
\caption{Results of BadCLIP with different maximum noise strengths. We report the harmonic mean of the results on the seen and unseen classes.}
\vspace{-1em}
\label{fig:effect_of_eps}
\end{figure}

\vspace{0.2em}
\noindent \textbf{Ablation on the maximum noise strength.}
% one table or one figure
\ In this part, we discuss the effect of the maximum noise strength on ACC and ASR. We set the parameter $\epsilon \in \{0.1,0.5,1,2,4\}$ and present the results in Fig. \ref{fig:effect_of_eps}. When $\epsilon$ is relatively small ($<$1), the ACC and ASR values increase with the increase of $\epsilon$. However, when $\epsilon$ is larger than 1, the performance of BadCLIP remains almost unchanged for different $\epsilon$. It illustrates that BadCLIP can be effective even with a very small noise strength. Considering the generalizability in various settings and the visual stealthiness described in Section 5.6, $\epsilon=4$ by default in our experiments is a reasonable choice.

\begin{table}[]
\caption{BadCLIP under defenses on Caltech101.}
\label{tab:defense}
\setlength{\tabcolsep}{1em}{
\resizebox{\linewidth}{!}{
\begin{tabular}{@{}lcc|cc|cc@{}}
\toprule
\multirow{2}{*}{Defense} & \multicolumn{2}{c|}{Seen} & \multicolumn{2}{c|}{Unseen} & \multicolumn{2}{c}{H} \\ \cmidrule(l){2-7} 
 & ACC & ASR & ACC & ASR & ACC & ASR \\ \midrule
N/A & 97.8 & 99.7 & 93.4 & 99.2 & 95.5 & 99.4 \\ \midrule
CleanCLIP & 97.7 & 89.2 & 95.2 & 86.6 & 96.4 & 87.9 \\
Fine-tuning & 97.5 & 98.9 & 95.3 & 99.2 & 96.4 & 99.0 \\
FT-SAM & 96.1 & 99.1 & 95.5 & 97.8 & 95.8 & 98.4 \\ \bottomrule
\end{tabular}}}
\end{table}

\section{Evaluation on More Defense Methods}
We evaluate BadCLIP on more defense methods, including CleanCLIP \cite{bansal2023cleanclip}, fine-tuning, and FT-SAM \cite{zhu2023enhancing}.
As shown in Table \ref{tab:defense}, BadCLIP still achieves high ASRs under these three defenses.  We also evlaute on the inference-time defense, TeCo \cite{liu2023detecting}. It results in a 0.55 AUROC, only slightly better than the random guess.
These results indicate that our attack is resistant to existing defenses.

\section{Limitation and Future Work} 
Despite promising performance of BadCLIP in most cases, there is a gap between the accuracy on clean images of BadCLIP and that of using hand-crafted prompts in unseen classes. In fact, this is a challenging problem for prompt learning methods \cite{zhou2022learning,zhou2022conditional,zhu2022prompt}, which is an interesting future direction for backdoor attacks on CLIP.  
Another limitation of BadCLIP is that it assumes that the attacker has full
knowledge of the pre-trained CLIP model including model architectures and parameters. We will
further explore more strict settings than the white-box one in our future work.

\end{document}